\documentclass[review,3p,times,sort&compress]{elsarticle}
\pdfoutput=1
\usepackage{lineno}
\modulolinenumbers[5]

\usepackage{hyperref}

\usepackage{caption}
\usepackage{subcaption}
\usepackage{amsmath}
\usepackage{lscape}
\usepackage{natbib}
\usepackage{graphicx}
\usepackage{multirow}
\usepackage{xspace}
\usepackage{pifont}% http://ctan.org/pkg/pifont
\newcommand{\cmark}{\ding{51}}%
\newcommand{\xmark}{\ding{55}}%
\let\oldCopyright\copyright
\usepackage[usenames,dvipsnames,svgnames,table]{xcolor}

\usepackage{todonotes}
\usepackage{sidecap}

\renewcommand{\copyright}{\textsuperscript{\oldCopyright}\xspace}

\xspaceaddexceptions{]\}*}
\newcommand{\latinphrase}[1]{\emph{#1}\@\xspace}
\newcommand{\ie}{\latinphrase{i.e.}}

\newcommand{\etal }{\latinphrase{et al.}}

\begin{document}

\begin{frontmatter}

\title{An Encoder-Decoder CNN  for Hair Removal in Dermoscopic Images}

\cortext[mycorrespondingauthor]{Corresponding author}

\author[addressuib,addressidisba]{Lidia Talavera-Mart{\'i}nez}\corref{mycorrespondingauthor}
\ead{l.talavera@uib.es}

\author[addressuib,addressidisba]{Pedro Bibiloni}
\ead{p.bibiloni@uib.es}
\author[addressuib,addressidisba]{ Manuel Gonz\'alez-Hidalgo}
\ead{manuel.gonzalez@uib.es}

\address[addressuib]{SCOPIA research group. University of the Balearic Islands. Dpt. of Mathematics and Computer Science. 
Crta. Valldemossa, km 7.5, E-07122 Palma, Spain}
\address[addressidisba]{Health Research Institute of the Balearic Islands (IdISBa), E-07010 Palma, Spain}

\begin{abstract}
The process of removing occluding hair has a relevant role in the early and accurate diagnosis of skin cancer. It consists on detecting hairs and restore the texture below them, which is sporadically occluded. In this work, we present a model based on convolutional neural networks for hair removal in dermoscopic images. During the network's training, we use a combined loss function to improve the restoration ability of the proposed model. In order to train the CNN and to quantitatively validate their performance, we simulate the presence of skin hair in hairless images extracted from publicly known datasets such as the PH2, dermquest, dermis, EDRA2002, and the ISIC Data Archive. As far as we know, there is no other hair removal method based on deep learning. Thus, we compare our results with six state-of-the-art algorithms based on traditional computer vision techniques by means of similarity measures that compare the reference hairless image and the one with hair simulated. Finally, a statistical test is used to compare the methods. Both qualitative and quantitative results demonstrate the effectiveness of our network.
\end{abstract}


\end{frontmatter}

\section{Introduction}

%\paragraph{Melanoma,Incidence,Dermoscopy,CAD}
Melanoma is the most aggressive, metastatic and deadliest type of skin cancers, turning this disease into a major problem for public health. In Europe, it accounts for 1--2\% of all malignant tumors \cite{mmmp}, and its estimated mortality in 2018 was 3.8 per 100.000 men and women per year \cite{ECIS}. Although melanoma is still incurable, its early diagnosis is of great importance. Its early detection can prevent malignancy and increase the survival rate and the effectiveness of the treatment. Nowadays, practitioners rely on the dermoscopic evaluation for completing the clinical analysis and the diagnosis of melanoma. This practice improves the diagnostic accuracy up to 10--30\% \cite{mayer1997systematic} compared to simple clinical observation. This in-vivo, non-invasive skin imaging technique enables the visualization of specific subsurface structures, forms and colors that are not discernible by a simple visual inspection. The diagnosis of skin lesions is mainly based on their morphological characteristics, such as an irregular shape, asymmetry and a variety of colors, along with a history of changes in size, shape, color and/or texture. However, their evaluation might be altered by the individual judgment of the observers, which depends on their experience and subjectivity \cite{argenziano2003dermoscopy}. Thus, in order to help physicians to obtain an early, objective, and reproducible diagnosis of skin lesions, sophisticated Computer-Aided Diagnosis (CAD) software are developed. These computational tools are designed based on clinical protocols \cite{argenziano1998epiluminescence,kittler2007dermatoscopy,menzies1996frequency,stolz1994abcd} and focus mainly on image acquisition, artifact removal (hairs, bubbles, etc.), segmentation of the lesion, extraction and selection of features, and final classification of the lesion.

%\paragraph{Motivacion}
In the pre-processing stage during the analysis of the lesions, hair removal is one of the key steps. The presence of hair in dermoscopic images usually occludes significant patterns reducing the accuracy of the system. Once the hairs are detected and removed, the next step is to estimate and restore the underlying information (i.e., color and texture patterns) of the skin pixels underneath the hair regions. Extensive previous research has been done addressing the hair removal process in dermoscopic images. To the best of our knowledge, previous works presented approaches based on traditional computer vision techniques. Thus, tackling the problem with classical generative and discriminative models, which rely on hand-crafted features. However, hand-crafted features needed to be defined and their impact is typically tested with small datasets \cite{o2019deep}. In recent years, deep learning has shown to be a powerful tool for image analysis. More specifically, deep learning techniques have achieved a higher performance with respect to traditional approaches for the majority of applications within the medical field \cite{litjens2017survey}. Deep Convolutional Neural Networks (CNNs) allow to automatically learn features of different complexity directly from data through a set of trainable filters. Moreover, it has shown to be a powerful tool when working with large datasets. 

% In this work, we propose a novel model for the task of hair removal in dermoscopic images. We rely on CNNs for the reconstruction of the image that implies the recognition of hairs and the later restoration of the underneath region.

%\paragraph{Objetivos}
As we have mentioned before, CAD systems may include a preprocessing step in which hair removal stands out as one of the most useful and used methods. However, traditional approaches are still used, nowadays, for this task in more advanced systems in which the main model uses deep learning techniques \cite{salido2018using,bakkouri2020computer}. Thus, we face the task of developing a deep learning model for the detection and removal of hairs. Such model could be integrated into a complete CAD system based entirely on deep learning. Our objective is threefold. First, to design a novel model based on deep CNNs for the removal of skin hair in dermoscopic images and the subsequent restoration of the affected pixels. Second, to qualitative and quantitatively assess its performance. Third, to compare the results with other hair removal strategies using the same database, which would provide an objective comparison of the strengths and weaknesses, and therefore of the quality of the method presented.

\iffalse{
\textcolor{red}{As we have mentioned before, CAD systems always have a preprocessing step in which hair removal stands out as one of the most useful and used methods. However, traditional approaches are still used, nowadays, for this task in more advanced systems in which the main model uses deep learning techniques \cite{salido2018using,bakkouri2020computer}. That is why the usefulness of the model presented in this work also resides in the possibility of integrating it into a complete CAD system based entirely on deep learning.}}\fi

In summary, we make the following contributions:
\begin{itemize}
    \item To the best of our knowledge, we are the first to propose the use of a CNN model for hair removal in dermoscopic images.
    \item We introduce a loss function for the detection and posterior restoration of hair’s pixels based on the combination of well-known loss functions. We also present an ablation study of the different loss terms.
    \item We extend the dataset presented in \cite{talavera2019comparative}, through the creation of more synthetic hair on dermoscopic images from public datasets. This allows us to carry out a quantitative evaluation of hair removal approaches.
\end{itemize}

%\paragraph{Description of the structure of the paper}
The rest of the document is structured as follows. First, in Section \ref{sec:relatedworks}, we review the related work on hair removal methods in dermoscopic images. Then, in Section \ref{sec:methodolooy}, we present our method based on CNNs and detail the loss function that we have used to train the network. Next, in Section \ref{sec:experimentsetup}, we describe the database and the implementation details, as well as presenting the results obtained, and performing an ablation study of the loss terms and some architecture's aspects. Finally, in Section \ref{sec:discussionresults}, we discuss the previous results and study the strengths and limitations of this study.

\section{Related Work} \label{sec:relatedworks}

This section briefly describes previous works that addressed the task of hair removal. 

Several traditional computer vision approaches have been used for hair removal in dermoscopic images. Here we highlight six state-of-the-art algorithms, based on their availability and wide use in the literature. These are the ones proposed by Lee \etal \cite{lee1997dullrazor}, Xie \etal \cite{xie2009pde}, Abbas \etal \cite{abbas2011hair}, Huang \etal \cite{huang2013robust}, Toossi \etal \cite{toossi2013effective} and Bibiloni \etal \cite{bibiloni2017skin}. A summary of the approaches considered by each of them can be seen in Table \ref{tab:reviewMethods}.
\begin{table}[h!]
\centering
\resizebox{12cm}{!} {
\begin{tabular}{lllll}
\hline
\multicolumn{1}{c}{Year} & \multicolumn{1}{c}{Method} & \multicolumn{1}{c}{Hair segmentation} & \multicolumn{1}{c}{Inpainting method} & \multicolumn{1}{c}{Color space} \\ \hline
1997 & Lee \etal \cite{lee1997dullrazor} & Grayscale closing & Bilinear interpolation & RGB \\ 
2009 & Xie \etal \cite{xie2009pde} & Grayscale top-hat & Non-linear PDE & RGB \\ 
2011 & Abbas \etal \cite{abbas2011hair} & Derivative of Gaussians & Coherence transport & CIELab \\ 
\multirow{2}{*}{2013} & \multirow{2}{*}{Huang \etal \cite{huang2013robust}} & \multirow{2}{*}{Conventional matched filters} & Region growing algorithms and & \multirow{2}{*}{RGB}\\ & & & linear discriminant analysis  \\
2013 & Toossi \etal \cite{toossi2013effective} & Canny edge detector & Coherence transport & RGB \\ 
2017 & Bibiloni \etal \cite{bibiloni2017skin} & Color top-hat & Morphological  inpainting & CIELab \\ \hline
\end{tabular}
}
\caption{Detection and inpainting techniques employed in the literature to remove hair from dermoscopic images.}
\label{tab:reviewMethods}
\end{table}

Deep learning techniques have been used to face a large number of tasks in computer vision. Specifically, deep learning-based image restoration techniques have been used in other fields for image inpainting \cite{xie2012image}, image deblurring and image denoising \cite{tian2019deep}, among others. These methods learn the parameters of the network to reconstruct images directly from training data, that is composed by pairs of clean and corrupted images. This is usually more effective in real-world images. For instance, Xie \etal \cite{xie2012image} proposed an approach for image denoising and blind inpainting that combines sparse coding with pre-trained deep networks, achieving good results in both tasks. Vincent \etal \cite{vincent2008extracting} presented a stack of denoising autoencoders for image denoising that is applied not only to the input, but also recursively to intermediate representations, to initialize the deep neural network. Also, Cui \etal \cite{cui2014deep} proposed a cascade of multiple stacked collaborative local autoencoders for image super-resolution. Their method searches in each layer non-local self-similarity to enhance high frequency texture details from the image patches to suppress the noise and combine the overlapping patches. In \cite{mao2016image}, Mao \etal proposed an encoding-decoding framework for image denoising and super-resolution combining convolutions and deconvolution layers linked symmetrically by skip connections, which helps improving the training process and the network’s performance. Finally, Jain \etal \cite{jain2009natural} and Dong \etal \cite{dong2015image} proposed a fully convolutional CNN for image denoising and image super-resolution, respectively. The authors shown that their methods achieve comparable results to traditional computer vision techniques.

Despite the good performance of the above mentioned works, and to the best of our knowledge, no previous research has been done on the field of hair removal in dermoscopic images relying on deep learning. Although Attia \etal works with this technique to simulate hair in dermoscopic images. This work aims to fill that gap and serve as a baseline for future works.
\section{Methodology}\label{sec:methodolooy}

In this section, we describe our model for hair removal in dermoscopic images. We specifically focus on our proposed reconstruction loss function.

%\subsection{Network Architecture}
We designed a convolutional encoder-decoder architecture for the task of hair removal in dermoscopic images. The model, with 12 layers, is described in Figure \ref{fig:architecture}.
The input of our model is the pair formed by the reference image without hair, used only when computing the loss function, and its corresponding image with simulated hair. On one hand, the encoder, with an input of size $512 \times 512 \times 3$, extracts a hidden representation of high-level features from the image. When looking for such features, the encoder tends to ignore noise. In our case, we expect the network to output hairless images, therefore hair becomes the noise that can be ignored by the encoder. On the other hand, the decoder aims to recover the missing information from the high-level feature representation. Its output is a $512 \times 512 \times 3$ cleaned version, without skin hair, of the input image. Both the encoder and the decoder have two blocks. Each block of the encoder consists of one $3\times3$ convolution, of 128 filters in the first block and 256 filters in the second one, followed by a down-sampling operation, which is applied by a two-strided $3\times3$ convolution to reduce the spatial resolution. On the other side, each block of the decoder consists of an up-sampling of the feature map by a deconvolution of $3\times3$ with strides of two in each dimension. A skip connection follows, which concatenates the up-sampled output with the corresponding feature map from the layer of equal resolution of the encoder. This enables the decoder to recover image details, and therefore improves the restoration performance.  Next, a $3\times3$ convolution is applied over the merged data. Finally, in the last block, an additional $3\times3$ convolution is added to reduce the feature map to the number of output channels.

\iffalse{
\begin{figure}
\includegraphics[height=240pt,width=\textwidth]{images/architecture.png}
\caption{Architecture of our proposed network. The pairs of reference hairless images and its corrupted (hair simulated) images are passed through the encoder to extract complex features while eliminates pixels of the hair regions. Finally, the decoder, which is connected to the encoder throughout skip-connections, produces the reconstructed image by inpainting the hair regions in the image.} \label{fig:architecture}
\end{figure}
}\fi

\begin{figure}
\includegraphics[height=280pt,width=\textwidth]{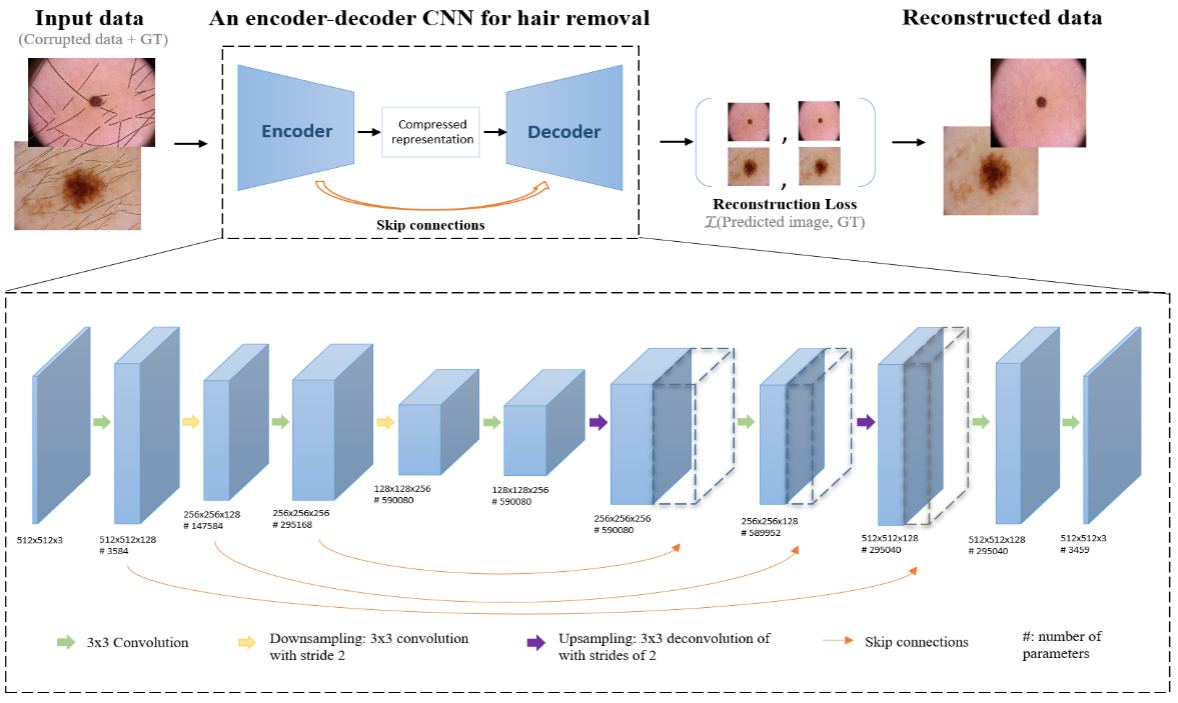}
\caption {Architecture of our proposed network. The pairs of reference hairless images and its corrupted (hair simulated) images are passed through the encoder to extract complex features. The decoder, connected to the encoder with skip-connections, reconstructs the image.} \label{fig:architecture}
\end{figure}

\subsection*{\textit{Reconstruction Loss Function}}

The loss function guides how the network learns by serving as a criterion that numerically reflects the errors of the model. It is computed between the network output and its corresponding hairless reference image, also known as Ground Truth (GT). There are several loss functions that have been used in image restoration tasks. Some widespread losses are the Mean Square Error (MSE) or the Mean average Error (MAE). These measurements exclusively depend on the difference between the corresponding pixels of the two images. Therefore, the results might have poor quality in terms of human perception, since the noise of a pixel should not be considered independently of the error of its neighbouring pixels. To overcome these limitations, other loss metrics have been proposed, such as the Structural Similarity Metric (SSIM) or the Multiscale Structural Similarity Metric (MSSSIM), which depend on local luminance, contrast and structure \cite{zhao2016loss}.

To achieve results appealing to a human observer, and inspired by the results obtained by Liu \etal in \cite{liu2018image}, we propose to capture the best characteristics of the loss functions that measure statistical features locally along with other per-pixel losses. 

\newcommand{\loss}[1]{\mathcal{L}_\text{#1}}

Thus, our reconstruction loss is defined as follows:
\begin{equation}
    % \loss{rec}=\alpha\loss{foreground}+\gamma\loss{background}+\beta\loss{SSIM}+\delta\loss{composed} + \lambda \loss{tv},
    \loss{rec}=\alpha L_1^\text{foreground}+\beta L_1^\text{background}+\gamma L_2^\text{composed}+\delta\loss{SSIM} + \lambda \loss{tv},
\end{equation}
where $\alpha, \beta, \gamma, \delta$ and $\lambda$ are the weights of each term of the linear combination that define the reconstruction loss function. We opted to perform a random hyperparameter search as there are many parameters of which we have to find the optimal value, and a grid search would require a higher computational cost. Specifically, we performed 10 runs of our model, assigning in each case a random value between 0 and 10 to each of the weights. Afterwards, a statistical test indicates which are the best set of weights according to the measurements explained in Section \ref{subsecquantitaive}. The term $L_1^\text{foreground}$ is the $L_1$ distance between the GT and the prediction of the network only between those pixels belonging to the hair areas. Next, $L_1^\text{background}$ estimates the $L_1$ distance between the GT and the network's prediction only among the background pixels. Then, $L_2^\text{composed}$ computes the $L_2$ restricted to the hair, but normalizing over all pixels rather than the amount of hair pixels. The term $\loss{SSIM}$ calculates the loss function based on the SSIM metric over the whole image. Finally, we use a total variation loss, $\loss{tv}$, as a regularizer to smooth the transition of the predicted values for the regions corresponding to hair, according to their surrounding context. A more detailed description of this term can be found in \cite{rudin1992nonlinear}.

\section{Experimental framework and Results}\label{sec:experimentsetup}
In this section, we first establish the experimental framework by describing the database used and the implementation details of our method. Then, we analyze the results obtained by our method and compare them, qualitative and quantitatively, to the six traditional hair removal methods presented in Section \ref{sec:relatedworks}. To obtain the numerical results, we rely on several performance measures. We determine which method outperforms the rest by means of a statistical test. Finally, we conducted an ablation study of the loss terms, as well as of some aspects of the model's architecture.

\subsection{Dataset description}

In order to train the CNN and to quantitatively validate, in an effective way, the performance of our method, we need a dataset. It must contain pairs of images: images with hair, used as the algorithm input, along with their corresponding ``clean" version, in this case the same image without hair.

Finding this type of data is a challenging task, the same dermoscopic image can not be captured with and without hair. To tackle this problem, we decided to simulate the presence of skin hair in hairless images extracted from five publicly available datasets, \ie PH2 \cite{mendonca2015ph2}, dermquest\footnote{Was deactivated on December 31, 2019}, dermis\footnote{\url{www.dermis.net}}, EDRA2002 \cite{argenziano2000interactive} and from the ISIC Data Archive\footnote{\url{www.isic-archive.com}}. We have avoided selecting images with other artifacts (eg. ruler, bandages etc) that are not hairs. Three different hair simulation methods have been used. The first is the one, presented by Attia \etal \cite{attia2020realistic}, is based on generative adversarial networks. The second one was implemented by Mirzaalian \etal \cite{mirzaalian2014hair}, whose software ``HairSim" is publicly available in \cite{hairsim}. Finally, the last approach we have used involved extracting hair masks by an automated method, proposed by Xie \etal \cite{xie2009pde}, and superimpose them on hairless images.

\iffalse{We constructed a dataset with \textcolor{red}{433} training images and \textcolor{red}{185} test images. \textcolor{red}{The training set consists of 270 images from the EDRA2002 dataset, 152 images from the PH2 dataset, 6 images from the dermis dataset, and 5 from the dermquest dataset. The test set has 46, 87, and 52 images from the ISIC Data Archive, PH2, and EDRA2002 datasets, respectively}.}\fi

\begin{figure}[!ht]
\centering
\resizebox{12.25cm}{!} {
\begin{tabular}{c}

    \includegraphics[height=90pt,width=0.3\textwidth]{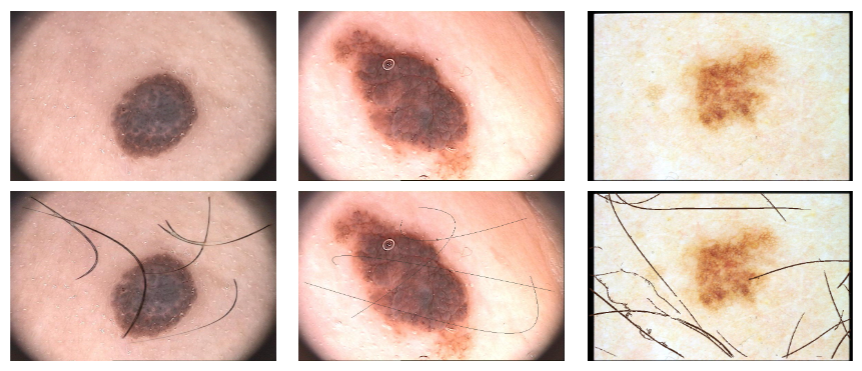}
\end{tabular}}
% \caption{Example of pairs of images with which the network trains. The images of the first row are the hairless images that act as GT. Then, each column of the second row represents one of the three different hair simulation methods used to generate the database. Example of hair simulation created by the deep neural network \cite{attia2018realistic} (bottom left), hair simulation created with the ``HairSim" software \cite{hairsim} (bottom in the middle), and hair simulation created by superimposing extracted hair mask (bottom right).}
\caption{
Original images (top) and simulated hair images (bottom), respectively by a deep neural network \protect\cite{attia2020realistic} (left), ``HairSim" \protect\cite{hairsim} (middle), superimposing a hair mask \protect\cite{xie2009pde} (right).
}
\label{fig:database}
\end{figure}

We constructed a dataset with 618 images, which consists of 322 images from the EDRA2002 dataset, 239 images from the PH2 dataset, 46 images from the ISIC Data Archive, 6 images from the dermis dataset, and 5 from the dermquest dataset. During the experimentation we divide it into 70\% for training and 30\% for the test, which gives us 433 and 185 images, respectively. It is composed of images with diverse hair that present a variety of hair thickness, density and color, ranging from coarse to more realistic. This diversity guarantees that we consider hairs with very different characteristics when training the network. We also introduced images without hair to teach the network that some images must not be modified and their textures must be maintained. In Figure \ref{fig:database}, we show examples of original hairless images from the dataset, together with the simulations obtained using each of the hair simulation approaches. As we can see, with the first and third method we achieve a much more natural morphology, quantity and distribution that resembles real hair compared to the second. We consider a reduced number of testing images to maximize the training samples and, thus, to help the network learn and generalize well.

\subsection{Experimental setup}\label{experimentalsetup}

We implemented the proposed architecture using Keras \cite{chollet2018keras}. The network was trained from scratch with randomly initialized weights and using the Adam \cite{kingma2014adam} optimizer with a learning rate experimentally set to $10^{-4}$. The coefficients for the different terms of the reconstruction loss function with which the network has been trained were experimentally found to be:
$\alpha=2.626$, $\beta=3.892$, $\gamma=0.309$, $\delta=0.398$ and $\lambda=0.597$. The network was trained on a single NVIDIA GeForce GTX 1070 with a batch size of 4.

\begin{figure}[h!]

    \includegraphics[width=\textwidth]{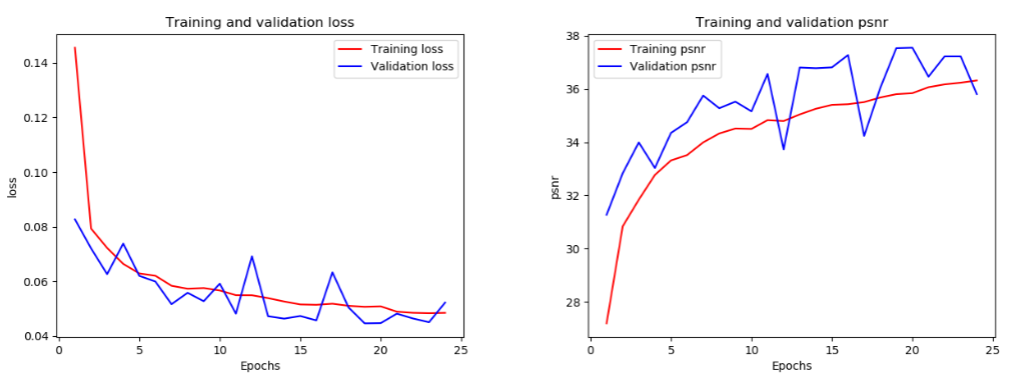}
    \caption{Plots of the Loss (left), and PSNR performance measure (right), of the training of the proposed model.}
    \label{fig:training}
\end{figure}

In figure \ref{fig:training}, we can see how our model trained over almost 25 epochs (reaching an early stopping policy based on monitoring the validation loss), and how the Peak Signal-to-Noise Ratio (PSNR) metric evolves satisfactory during the training.

\subsection{Qualitative results}
\begin{figure}[h!]
    
    \includegraphics[width=\textwidth]{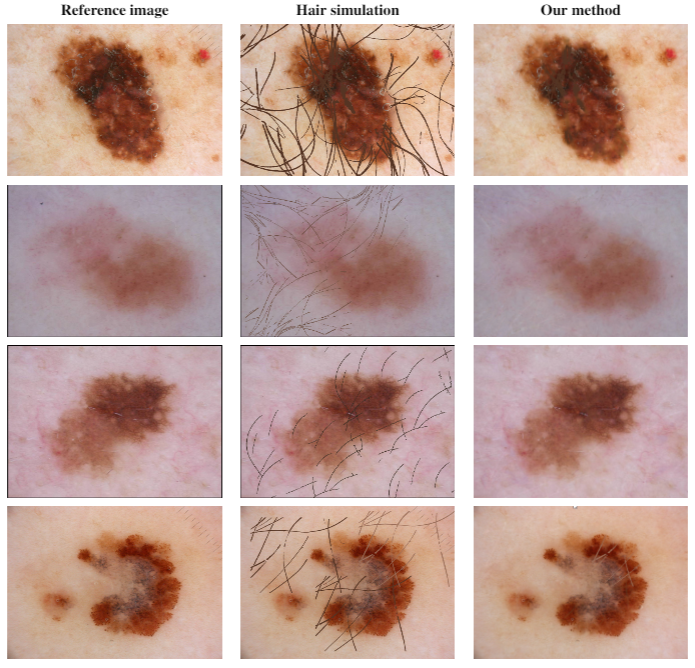}
    
    \caption{Example of the hair removal results obtained by our method.}
    \label{fig:graphics_results}
\end{figure}

We conducted a qualitative and quantitative analysis of the results obtained. In Figure \ref{fig:graphics_results}, we can see that our proposed method attains visually appealing results. In addition, we present a visual comparison of our results with the with the ones obtained by other hair removal methods presented in Section \ref{sec:relatedworks}. Let us remark that we compare with these methods as they are the ones that have been applied to dermoscopic images.

\begin{figure}[h!]

    \includegraphics[width=\textwidth]{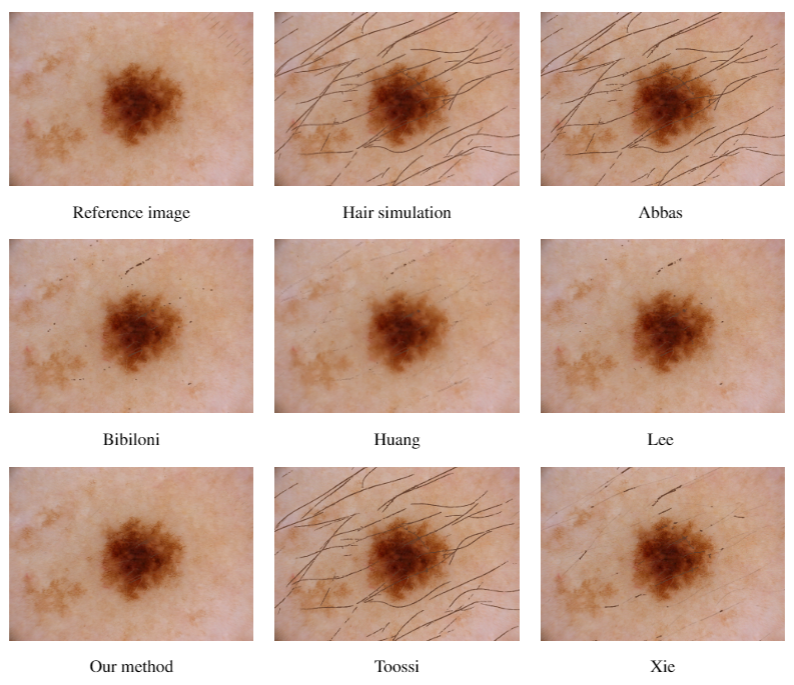}

    \caption{Example of the hair removal results obtained by the different algorithms.}
    \label{fig:graphic_comparation}
\end{figure}

As can be seen in Figure \ref{fig:graphic_comparation}, not all the methods are successful in both the hair removal and the subsequent process of inpainting. For instance, Abbas \etal's and Toossi \etal's methods are not capable of segment the hairs as it seems they are not able to detect them properly. In contrast, Huang \etal's and Xie \etal's methods are capable of detecting much of the hair. However, their inpainting process seems to leave a trail of them. Finally, our results and the ones of Bibiloni \etal's and Lee \etal's methods seem to adjust to the reference image at firs sight. Although, the last two leave traces, while the new method does not. However, it may be the case that some of them introduce some alterations, to a greater or lesser extent, that blur the lesion's features, such as streaks or reticular textures.

\begin{figure}[!ht]
     \centering
     \begin{subfigure}[b]{\textwidth}
         \centering
         \resizebox{0.4\linewidth}{!}{
         \includegraphics[width=\textwidth]{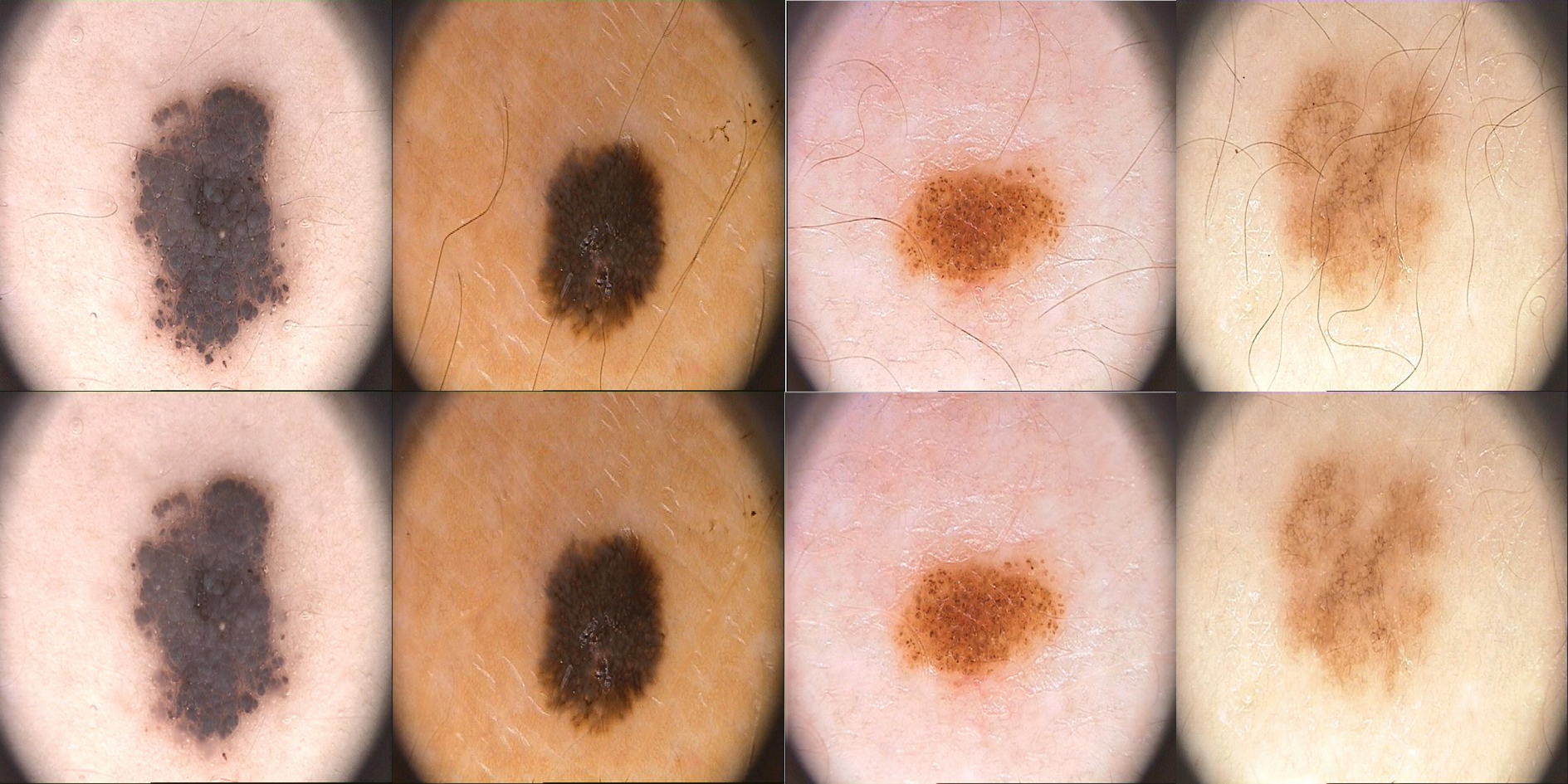}}
         \subcaption{}
     \end{subfigure}
     \\
     \begin{subfigure}[b]{\textwidth}
         \centering
         \resizebox{0.4\linewidth}{!}{
         \includegraphics[width=\textwidth]{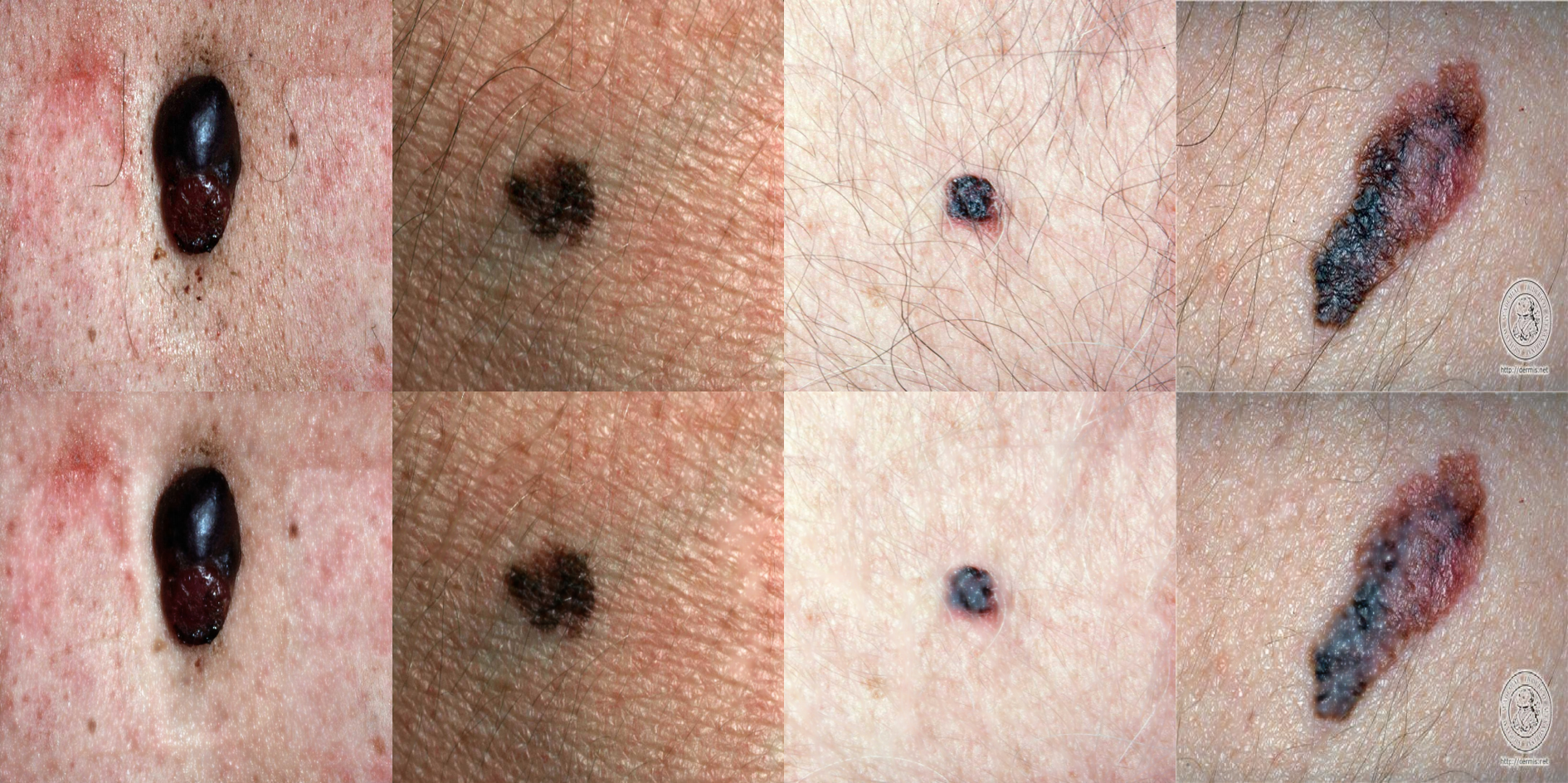}}
         \subcaption{}
     \end{subfigure}
     \\
     \begin{subfigure}[b]{\textwidth}
         \centering
         \resizebox{0.4\linewidth}{!}{
         \includegraphics[width=\textwidth]{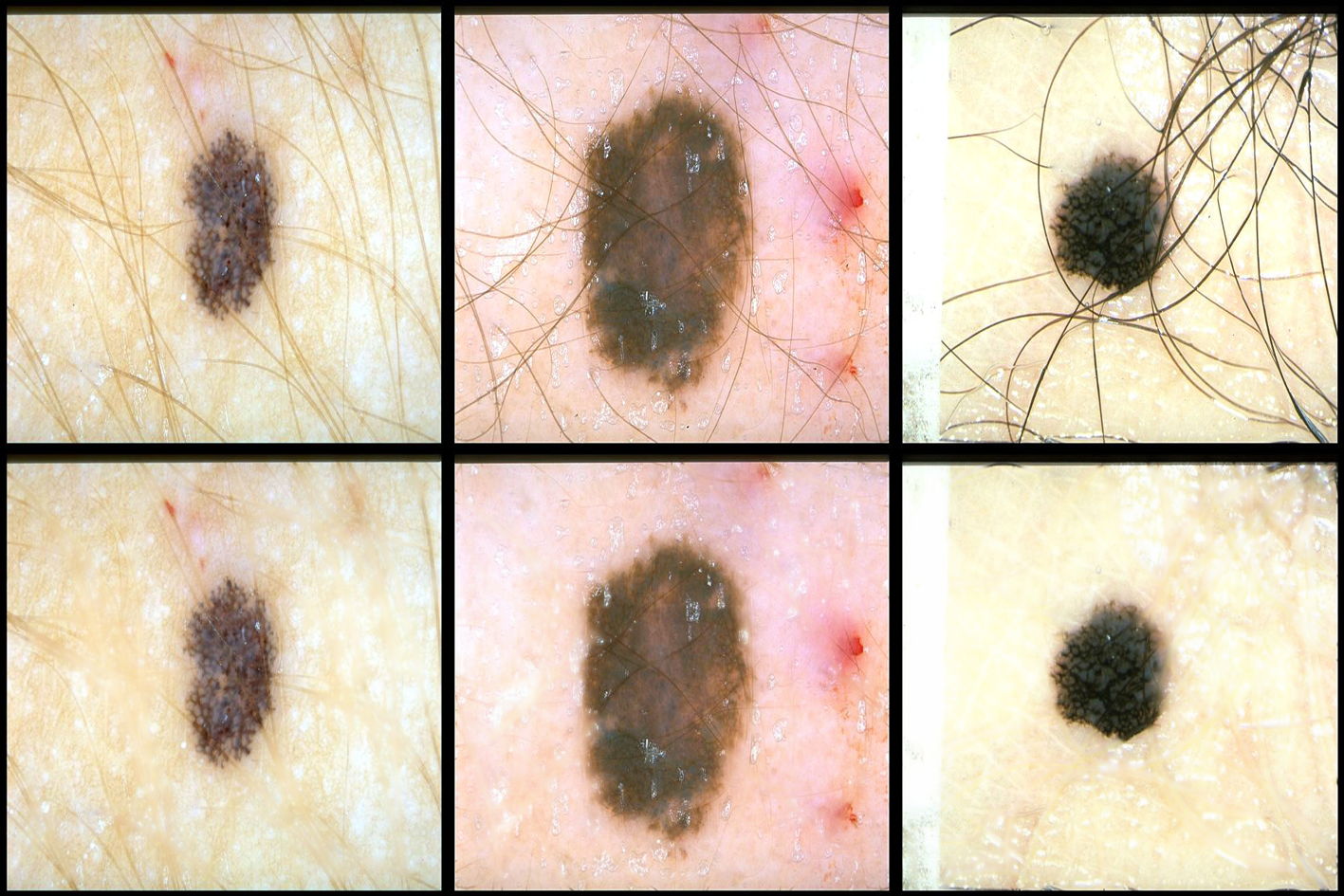}}
         \subcaption{}
     \end{subfigure}
     \\
    \begin{subfigure}[b]{\textwidth}
         \centering
         \resizebox{0.4\linewidth}{!}{
         \includegraphics[width=\textwidth]{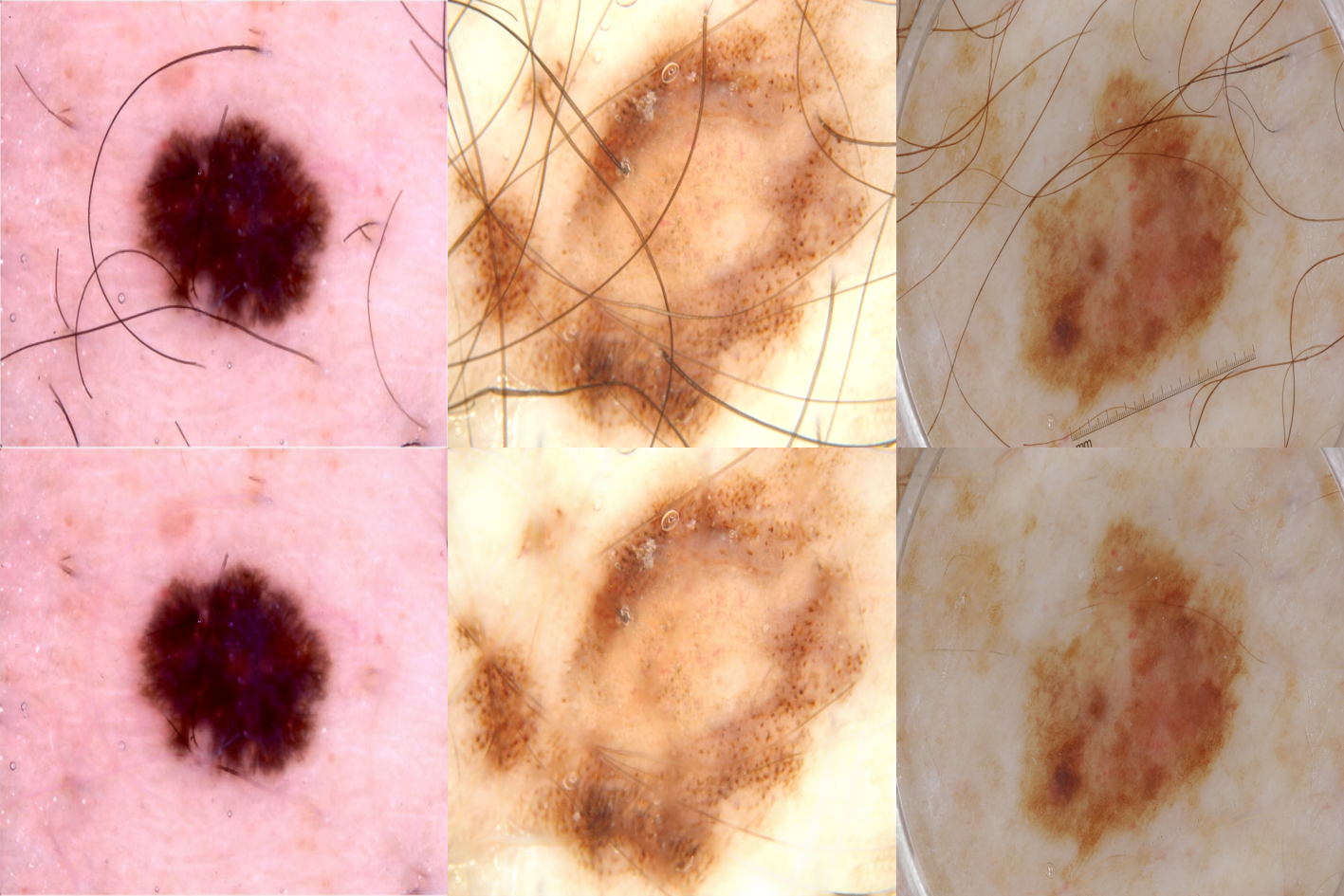}}
         \caption{}
     \end{subfigure}

        \caption{Example of hair removal results obtained by our method in dermoscopic images with real hair from (a) PH2 dataset, (b) Dermis and Dermquest datasets, (c) EDRA2002 dataset, and (d) ISIC Data Archive.}
        \label{fig:real_hair}
\end{figure}

Both in Figure \ref{fig:graphics_results} and \ref{fig:graphic_comparation}, we have seen that our method reaches good visual results when evaluated on synthetic images. In Figure \ref{fig:real_hair}, we show its effectiveness and its generalization ability in dermoscopic images with real hair. We show images from the 5 databases, to demonstrate that although the data is not balanced, the network has not suffered database-specific overfitting.

\subsection{Quantitative results}\label{subsecquantitaive}
It is worth noting that once the hairs are removed there are many possible solutions as to what is the expected inpainting result, always with the aim of preserving the texture of the area involved. Therefore, a qualitative evaluation is not enough to evaluate the quality of the different methods introduced. In the following, we introduce an automatic, objective and comparable performance evaluation system.

We used a set of nine objective error metrics to quantitatively assess the quality of the results obtained by the proposed CNN-based hair removal approach, with respect to the original hairless image. We cluster these measures into three different groups. The first one are the Mean Squared Error (MSE) \cite{wang2009mean}, the Peak Signal-to-Noise Ratio (PSNR) \cite{wang2004image}, the Root Mean Squared Error (RMSE) \cite{barnston1992correspondence}, and the Structural Similarity Index (SSIM) \cite{wang2004image}, which are per-pixel metrics. Within the second group we consider the Multi-Scale Structural Similarity Index (MSSSIM) \cite{wang2003multiscale} and the Universal Quality Image Index (UQI) \cite{wang2002universal}, which measure statistical features locally and then combine them. Finally, the Visual Information Fidelity (VIF) \cite{sheikh2004image}, the PSNR-HVS-M \cite{egiazarian2006new} and PSNR-HVS \cite{egiazarian2006new}, conforming the third group, have been designed to obtain more similar results to those perceived by the Human Visual System (HVS). This set of metrics constitutes a representative selection of the state-of-the-art performance metrics for restoration quality. We must recall that largest values of PSNR, SSIM, MSSSIM, UQI, VIF,  PSNR-HVS-M, and PSNR-HVS are indicators of a better quality of reconstructed images. On the other hand, lowest values of MSE and RMSE are indicators of higher similarity.

In Table \ref{tab:metrics}, we show the mean and standard deviation of the results obtained for the 185 images of the test set, and for each of the nine performance measures. In addition, we make a comparison of our results against the ones obtained by applying the six state of-the-art hair removal methods, detailed in Section \ref{sec:relatedworks}, to the same 185 images.

\begin{table}[h!]
\centering
\resizebox{12.25cm}{!} {
\begin{tabular}{lcrrrrrrr}
%\cline{3-8} 
 & \multicolumn{1}{l}{} &
 \multicolumn{1}{c}{Our method} &\multicolumn{1}{c}{Abbas} & \multicolumn{1}{c}{Bibiloni}  & \multicolumn{1}{c}{Huang} & \multicolumn{1}{c}{Lee} &  \multicolumn{1}{c}{Toossi} & \multicolumn{1}{c}{Xie} \\ \hline \hline
\multirow{2}{*}{MSE} & $\mu$ & 27.847 & 258.347 & 103.903 & 404.366 & 175.303 &  221.346 & 55.311 \\ \cline{2-9} 
 &$\sigma$& 35.261 & 302.243 & 108.762 & 589.341 & 219.633 &  256.319 & 103.553 \\ \hline
\multirow{2}{*}{SSIM} & $\mu$ & 0.926  & 0.867 & 0.885 & 0.851 & 0.890 &  0.864 & 0.921 \\ \cline{2-9} 
 &$\sigma$& 0.026 & 0.051 & 0.048 & 0.050 & 0.075 &  0.053 & 0.041 \\ \hline
\multirow{2}{*}{PSNR} & $\mu$ & 35.137 & 26.080 & 29.785 & 27.001 & 29.158 &  26.570 & 33.096 \\ \cline{2-9} 
 &$\sigma$& 3.006 & 4.079 & 4.073 & 6.603 & 5.783 &  3.852 & 3.705 \\ \hline
\multirow{2}{*}{RMSE} & $\mu$ & 4.790 & 14.226 & 9.207 & 15.485 & 11.032 &  13.287 & 6.318 \\ \cline{2-9} 
 &$\sigma$& 2.220  & 7.501 & 4.386 & 12.864 & 7.340 &  6.711 & 3.935 \\ \hline
\multirow{2}{*}{VIF} & $\mu$ & 	0.525  & 0.526 & 0.499 & 0.402 & 0.531 &  0.509 & 0.592 \\ \cline{2-9} 
 &$\sigma$& 0.099  & 0.180 & 0.176 & 0.130 & 0.186 &  0.178 & 0.186 \\ \hline
\multirow{2}{*}{UQI} & $\mu$ & 0.997 & 0.991 & 0.996 & 0.990 & 0.995 &  0.992 & 0.997 \\ \cline{2-9} 
 &$\sigma$& 0.004 & 0.009 & 0.005 & 0.013 & 0.006 &  0.008 & 0.004 \\ \hline
\multirow{2}{*}{MSSSIM} & $\mu$ & 0.978 & 0.870 & 0.934 & 0.917 & 0.945 &  0.875 & 0.955 \\ \cline{2-9} 
 &$\sigma$& 0.011 & 0.073 & 0.039 & 0.042 & 0.051 &  0.069 & 0.037 \\ \hline
\multirow{2}{*}{PSNR-HVS-M} & $\mu$ & 36.802 & 25.078 & 29.404 & 26.248 & 28.445 &  25.519 & 33.005 \\ \cline{2-9} 
 &$\sigma$& 3.703 & 3.861 & 4.245 & 6.854 & 6.247 &  3.680 & 4.106 \\ \hline
\multirow{2}{*}{PSNR-HVS} & $\mu$ & 35.168 & 24.628 & 28.681 & 25.738 & 27.826 &  25.065 & 32.186 \\ \cline{2-9} 
 &$\sigma$& 3.465 & 3.823 & 4.158 & 6.634 & 5.991 &  3.639 & 3.896 \\ \hline
\end{tabular}
}
\caption{Mean and standard deviation of the similarity measures obtained to compare our method with six state of the art hair removal algorithms.}
\label{tab:metrics}
\end{table}

%\subsubsection{Statistical study}
The next step in our work, is to study if one algorithm outperforms another one significantly. Given a fixed similarity measure, we decided to use a statistical test to contrast the means of all pairs of algorithms. Specifically, we have used the $t$-test if the samples pass the Shapiro-Wilk normality test, or the Wilcoxon signed-rank test, otherwise, both considering a significance level of 0.05. According to the statistical test we can determine which method surpasses the others. Table \ref{tab:statistic_test} summarizes the results obtained. In it, rows represent all the pairs of algorithms in which the statistical test was applied, and the columns correspond to the measures of similarity. As an example, let us interpret the test comparing \textit{Abbas vs Huang}. According to the SSIM and VIF performance measures, Abbas' algorithm significantly outperforms Huang's algorithm. While according to the PSNR, MSSIM, PSNR-HVS-M and PSNR-HVS measures, Huang's algorithm significantly outperforms Abbas' algorithm. For the rest of measures, both methods obtain statistically comparable results. Let us remark that Abbas' algorithm is non-statistically superior in all of them.

\begin{table}[h!]
\centering
\resizebox{15cm}{!} {
\begin{tabular}{llccccccccc}
\multicolumn{1}{c}{\textbf{}} &  & MSE & SSIM & PSNR & RMSE & VIF & UQI & MSSSIM & PSNR-HVS-M & PSNR-HVS \\ \hline \hline
\multirow{2}{*}{Our method \emph{vs.} Abbas} & $p$-value & 2.62e-28 & 1.51e-29 & 1.49e-30 & 3.89e-29 & 0.452 & 1.09e-28 & 4.49e-32 & 1.87e-31 & 3.33e-31 \\
 & Statistical test & \cmark\cmark & \cmark\cmark & \cmark\cmark & \cmark\cmark & \xmark & \cmark\cmark & \cmark\cmark  & \cmark\cmark & \cmark\cmark \\ \hline
\multirow{2}{*}{Our method \emph{vs.} Bibiloni} & $p$-value & 4.44e-30 & 1.29e-25 & 3.38e-31 & 1.27e-30 & 4.07e-04 & 2.05e-10 & 2.68e-30 & 1.06e-31 & 1.49e-31 \\
 & Statistical test & \cmark\cmark & \cmark\cmark & \cmark\cmark & \cmark\cmark & \cmark\cmark & \cmark\cmark & \cmark\cmark & \cmark\cmark & \cmark\cmark \\ \hline
\multirow{2}{*}{Our method \emph{vs.} Lee} & $p$-value & 1.21e-30 & 2.23e-17 & 4.66e-30 & 1.57e-30 & 0.160 & 8.52e-13 & 3.35e-27 & 1.81e-31 & 3.85e-31 \\
 & Statistical test & \cmark\cmark & \cmark\cmark & \cmark\cmark & \cmark\cmark & \xmark & \cmark\cmark & \cmark\cmark  & \cmark\cmark & \cmark\cmark \\ \hline
\multirow{2}{*}{Our method \emph{vs.} Huang} & $p$-value & 5.54e-30 & 2.18e-30 & 1.75e-31 & 9.41e-31 & 1.46e-23 & 1.51e-29 & 1.87e-31 & 7.19e-32 & 9.63e-32 \\
 & Statistical test & \cmark\cmark & \cmark\cmark & \cmark\cmark & \cmark\cmark & \cmark\cmark & \cmark\cmark & \cmark\cmark  & \cmark\cmark & \cmark\cmark \\ \hline
\multirow{2}{*}{Our method \emph{vs.} Toossi} & $p$-value & 1.49e-27 & 3.95e-29 & 5.12e-30 & 1.60e-28 & 1.49e-03 & 7.83e-27 & 5.28e-32 & 2.79e-31 & 5.83e-31 \\
 & Statistical test & \cmark\cmark & \cmark\cmark & \cmark\cmark & \cmark\cmark & \cmark\cmark & \cmark\cmark & \cmark\cmark & \cmark\cmark & \cmark\cmark \\ \hline
\multirow{2}{*}{Our method \emph{vs.} Xie} & $p$-value & 1.86e-14 & 9.59e-04 & 1.14e-14 & 1.32e-14 & 2.31e-10 & 4.05e-05 & 1.36e-25 & 1.72e-19 & 7.22e-18 \\
 & Statistical test & \cmark\cmark & \cmark\cmark & \cmark\cmark & \cmark\cmark & \cmark\cmark & \cmark\cmark & \cmark\cmark & \cmark\cmark & \cmark\cmark\\ \hline
\multirow{2}{*}{Abbas \emph{vs.} Bibiloni} & $p$-value & 5.24e-23 & 1.27e-12 & 1.40e-23 & 2.43e-23 & 0.012 & 8.04e-29 & 3.07e-27 & 1.39e-24 & 3.42e-24 \\
 & Statistical test & \xmark\xmark & \xmark\xmark & \xmark\xmark & \xmark\xmark & \cmark\cmark & \xmark\xmark & \xmark\xmark & \xmark\xmark & \xmark\xmark \\ \hline
\multirow{2}{*}{Abbas \emph{vs.} Lee} & $p$-value & 1.36e-11 & 4.70e-16 & 2.48e-15 & 7.87e-13 & 1.03e-10 & 5.69e-23 & 3.96e-26 & 1.63e-15 & 1.22e-15 \\
 & Statistical test & \xmark\xmark & \xmark\xmark & \xmark\xmark & \xmark\xmark & \xmark\xmark & \xmark\xmark & \xmark\xmark & \xmark\xmark & \xmark\xmark \\ \hline
\multirow{2}{*}{Abbas \emph{vs.} Huang} & $p$-value & 0.839 & 1.22e-03 & 5.1e-02 & 0.488 & 3.95-e21 & 0.118 & 1.49e-17 & 1.61e-03 & 1.56e-03 \\
 & Statistical test & \cmark & \cmark\cmark & \xmark\xmark & \cmark & \cmark\cmark & \cmark & \xmark\xmark & \xmark\xmark & \xmark\xmark \\ \hline
\multirow{2}{*}{Abbas \emph{vs.} Toossi} & $p$-value & 9.08e-28 & 3.52e-07 & 2.93e-27 & 1.27e-27 & 4.14e-32 & 4.76e-25 & 3.27e-08 & 5.66e-25 & 4.13e-25 \\
 &Statistical test & \xmark\xmark  & \cmark\cmark & \xmark\xmark & \xmark\xmark & \cmark\cmark & \xmark\xmark & \xmark\xmark & \xmark\xmark & \xmark\xmark \\ \hline
\multirow{2}{*}{Abbas \emph{vs.} Xie} & $p$-value & 4.56e-32 & 4.21e-32 & 4.64e-32 & 4.56e-32 & 4.64e-32 & 8.74e-32 & 5.64e-32 & 4.87e-32 & 4.87e-32 \\
 & Statistical test & \xmark\xmark & \xmark\xmark & \xmark\xmark & \xmark\xmark & \xmark\xmark & \xmark\xmark & \xmark\xmark & \xmark\xmark & \xmark\xmark \\ \hline
\multirow{2}{*}{Bibiloni \emph{vs.} Lee} & $p$-value & 5.07e-05 & 6.25e-13 & 2.51e-02 & 3.71e-04 & 1.34e-17 & 2.05e-10 & 1.36e-15 & 2.42e-03 & 2.36e-03 \\
 & Statistical test & \cmark\cmark & \xmark\xmark & \cmark\cmark & \cmark\cmark & \xmark\xmark & \cmark & \xmark\xmark & \cmark\cmark & \cmark\cmark \\ \hline
\multirow{2}{*}{Bibiloni \emph{vs.} Huang} & $p$-value & 1.53e-13 & 9.88e-24 & 9.07e-15 & 7.50e-14 & 4.51e-17 & 1.72e-17 & 4.58e-14 & 2.30e-16 & 3.38e-16 \\
 & Statistical test & \cmark\cmark & \cmark\cmark & \cmark\cmark & \cmark\cmark & \cmark\cmark & \cmark\cmark & \cmark\cmark & \cmark\cmark & \cmark\cmark \\ \hline
\multirow{2}{*}{Bibiloni \emph{vs.} Toossi} & $p$-value & 7.98e-21 & 1.59e-14 & 2.37e-21 & 3.29e-21 & 0.298 & 1.56e-27 & 4.32e-27 & 5.77e-23 & 1.22e-22 \\
 & Statistical test & \cmark\cmark & \cmark\cmark & \cmark\cmark & \cmark\cmark & \xmark & \cmark\cmark & \cmark\cmark & \cmark\cmark & \cmark\cmark \\ \hline
\multirow{2}{*}{Bibiloni \emph{vs.} Xie} & $p$-value & 1.03e-18 & 4.40e-29 & 2.24e-20 & 1.66e-19 & 4.21e-32 & 5.35e-12 & 8.77e-14 & 1.74e-18 & 1.68e-19 \\
 & Statistical test & \xmark\xmark & \xmark\xmark & \xmark\xmark & \xmark\xmark & \xmark\xmark & \xmark\xmark & \xmark\xmark & \xmark\xmark & \xmark\xmark \\ \hline
\multirow{2}{*}{Lee \emph{vs.} Huang} & $p$-value & 1.24e-11 & 6.62e-17 & 1.61e-15 & 9.23e-13 & 4.76e-20 & 2.92e-21 & 1.44e-17 & 9.01e-16 & 1.92e-15 \\
 & Statistical test & \cmark\cmark & \cmark\cmark & \cmark\cmark & \cmark\cmark & \cmark\cmark & \cmark\cmark & \cmark\cmark & \cmark\cmark & \cmark\cmark \\ \hline
\multirow{2}{*}{Lee \emph{vs.} Toossi} & $p$-value & 4.13e-07 & 1.75e-16 & 9.88e-12 & 7.17e-09 & 2.16e-14 & 1.32e-20 & 4.08e-26 & 4.36e-12 & 3.29e-12 \\
 & Statistical test & \cmark\cmark & \cmark\cmark & \cmark\cmark & \cmark\cmark & \cmark\cmark  & \cmark\cmark & \cmark\cmark & \cmark\cmark & \cmark\cmark \\ \hline
\multirow{2}{*}{Lee \emph{vs.} Xie} & $p$-value & 1.11e-16 & 2.09e-10 & 1.15e-15 & 1.63e-16 & 6.04e-28 & 5.12e-06 & 0.869 & 2.22e-16 & 8.65e-17 \\
 & Statistical test & \xmark\xmark & \xmark\xmark & \xmark\xmark & \xmark\xmark & \xmark\xmark & \xmark\xmark & \xmark & \xmark\xmark & \xmark\xmark \\ \hline
\multirow{2}{*}{Huang \emph{vs.} Toossi} & $p$-value & 0.496 & 7.83e-03 & 0.133 & 0.871 & 1.70e-17 & 0.594 & 5.44e-16 & 0.050 & 5.38e-02 \\
 & Statistical test &\xmark  & \xmark\xmark & \cmark & \xmark & \xmark\xmark & \xmark & \cmark\cmark & \cmark\cmark & \cmark \\ \hline
\multirow{2}{*}{Huang \emph{vs.} Xie} & $p$-value & 6.91e-25 & 7.80e-32 & 9.36e-27 & 8.70e-26 & 2.82e-28 & 2.40e-24 & 1.08e-25 & 7.17e-27 & 3.89e-27 \\
 & Statistical test & \xmark\xmark & \xmark\xmark & \xmark\xmark & \xmark\xmark & \xmark\xmark & \xmark\xmark & \xmark\xmark & \xmark\xmark & \xmark\xmark \\ \hline
\multirow{2}{*}{Toossi \emph{vs.} Xie} & $p$-value & 4.87e-32 & 4.27e-32 & 5.11e-32 & 4.87e-32 & 4.14e-32 & 1.35e-31 & 7.55e-32 & 1.72e-19 & 5.11e-32 \\
 & Statistical test & \xmark\xmark & \xmark\xmark & \xmark\xmark & \xmark\xmark & \xmark\xmark & \xmark\xmark & \xmark\xmark & \xmark\xmark & \xmark\xmark \\ \hline
\end{tabular}
}
\caption{\small Classification of algorithms according to objective similarity measures. The results are as follows: \cmark\cmark \, if the population mean of the first algorithm is better than that of the second algorithm; 
\cmark \, if the mean of the first algorithm is better but statistically comparable to that of the second algorithm; 
\xmark \, if the mean of the first algorithm is worse but statistically comparable to that of the second algorithm; 
\xmark\xmark \, if the population mean of the first algorithm is worse than that of the second algorithm.}
\label{tab:statistic_test}
\end{table}

As it can be seen in Table \ref{tab:statistic_test}, taking into account all the considered performance measures, the proposed method out-stands according to the majority of similarity measures. It is only significantly outperformed on the VIF performance measure compared to Abbas \etal' and Lee \etal's algorithms. Among the rest of methods, we can see that Lee \etal's algorithm surpasses statistically Huang \etal' and Toossi \etal's algorithms. However, when comparing Lee \etal's algorithm with Bibiloni \etal's, the former is statistically better in very specific settings, namely the SSIM, VIF and  MSSSIM measures. It is Xie \etal's algorithm that outperforms Lee \etal's in all performance measures. In the comparison between the algorithms of Bibiloni with Huang and Toossi, it is the first that outperforms statistically the other two in the majority of measures. Finally, the Toossi \etal's algorithm is statistically superior to the Abbas \etal's, except in the SSIM and VIF performance measures. These two algorithms are the ones that provide statistically worse results compared to the rest of the algorithms.

\subsection{Ablation study}

\begin{figure}[h!]

    \includegraphics[width=\textwidth]{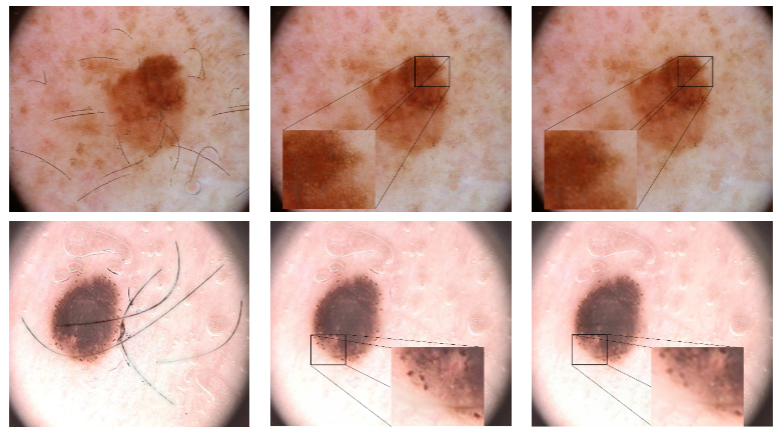}

    \caption{Original images (left), result with skip-connections (middle), and without skip-connections (right).}
    \label{fig:skipCon_ablation}
\end{figure}

Some works \cite{mao2016image, liu2018image} defend the fact that using skip connections, or convolutions/deconvolutions instead of pooling/unpooling layers may decrease the amount of detail loss and deteriorate the restoration performance. We study how these layers can affect the learning of our model by replacing the pooling layers with convolutions, and introducing skip-connection layers. In Table \ref{tab:DeConvSkipvs.PoolNoSkip}, we can see that the introduction of skip-connections does improve the results numerically, in terms of the similarity measures previously presented. However, these do not vary significantly when pooling layers are used instead of convolutional ones in our model. In this case, we can visualize its effects in Figure \ref{fig:skipCon_ablation}, where we compare the results of using or not skip-connections. As can be seen, the network is able to create a more detailed prediction with them, especially when it comes to dermoscopic structures such as streaks or globules.

\begin{table}[h!]
\centering
\resizebox{16cm}{!} {
\begin{tabular}{lrrrrrrrrr}
 & \multicolumn{1}{c}{MSE} & \multicolumn{1}{c}{SSIM} & \multicolumn{1}{c}{PSNR} & \multicolumn{1}{c}{RMSE} & \multicolumn{1}{c}{VIF} & \multicolumn{1}{c}{UQI} & \multicolumn{1}{c}{MSSSIM} & \multicolumn{1}{c}{PSNR-HVS-M} & \multicolumn{1}{c}{PSNR-HVS} \\ \hline \hline
\begin{tabular}[c]{@{}l@{}}Full method\end{tabular} & 27.847 & 0.926 & 35.137 & 4.790 & 0.525 & 0.997 & 0.978 & 36.802 &  35.168
 \\ \hline
\begin{tabular}[c]{@{}l@{}}Model with Pooling Layers\\ and Skip-Connections\end{tabular} & 27.786 & 0.925 & 35.122 & 4.794 & 0.524 & 0.998 & 0.978 & 36.596 & 35.003  \\ \hline
\begin{tabular}[c]{@{}l@{}}Model with DeConv Layers\\ and no Skip-Connections\end{tabular} & 32.846 & 0.909 & 33.862 & 5.443 & 0.467 & 0.997 &	0.974 &	35.538 & 33.908
  \\ \hline
\begin{tabular}[c]{@{}l@{}} Loss with no $L_1^\text{foreground}$ term \end{tabular} & 41.197 & 0.881 & 32.719 & 6.138 & 0.437 & 0.997 & 0.954 & 33.946 & 32.717
  \\ \hline
\begin{tabular}[c]{@{}l@{}} Loss with no $L_1^\text{background}$ term\end{tabular} & 26.398 & 0.928 & 35.306 & 4.689 & 0.521 & 0.997 & 0.979 & 36.862 & 35.209
\\ \hline
\begin{tabular}[c]{@{}l@{}} Loss with no $\loss{SSIM}$ term\end{tabular} & 31.220 & 0.926 & 34.488 & 5.119 & 0.523 & 0.997 & 0.979 & 36.802 & 35.154
  \\ \hline
\begin{tabular}[c]{@{}l@{}} Loss with no $L_2^\text{composed}$ term \end{tabular} & 30.881 & 0.926 & 34.463 & 5.113 & 0.523 & 0.997 & 0.978 & 36.744 & 35.143
  \\ \hline
\begin{tabular}[c]{@{}l@{}} Loss with no  $\loss{tv}$ term \end{tabular} & 30.621 & 0.923 & 34.673 & 5.039 & 0.522 & 0.997 & 0.977 & 36.599 & 34.985
  \\ \hline
\end{tabular}
}
\caption{Mean of the similarity measures obtained on the test set for the ablation study. The second and third row correspond to the skip-connections and pooling layers' ablation study. Then, from the fourth to the last row correspond to the ablation study of the loss terms.}
\label{tab:DeConvSkipvs.PoolNoSkip}
\end{table}

\begin{figure}[htpb!]
    \centering
     \begin{subfigure}[b]{0.3\textwidth}
         \centering
         \includegraphics[width=\textwidth]{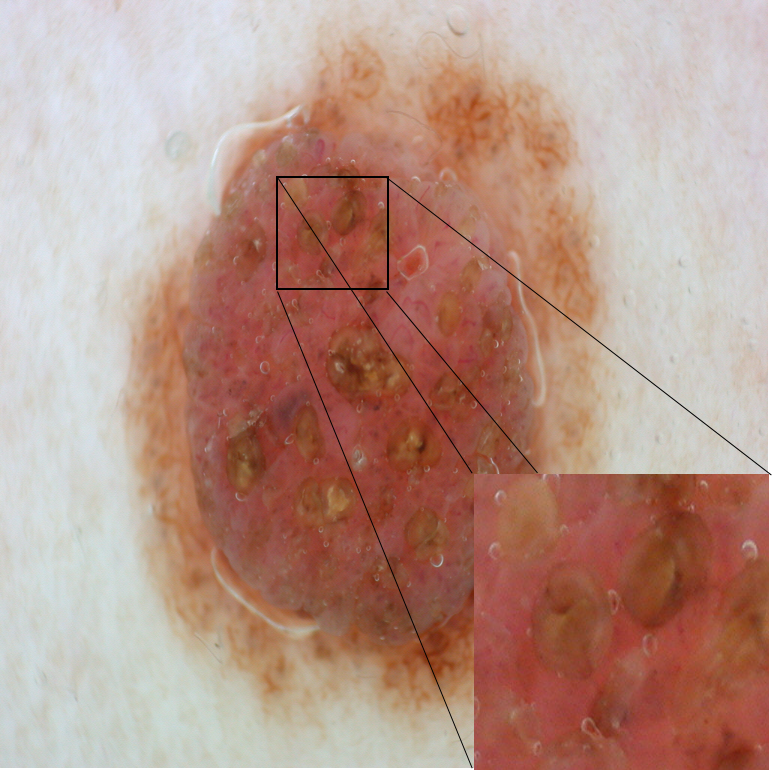}
         \caption{}
         \label{fig:ablationLoss_original}
     \end{subfigure}
     \begin{subfigure}[b]{0.3\textwidth}
         \centering
         \includegraphics[width=\textwidth]{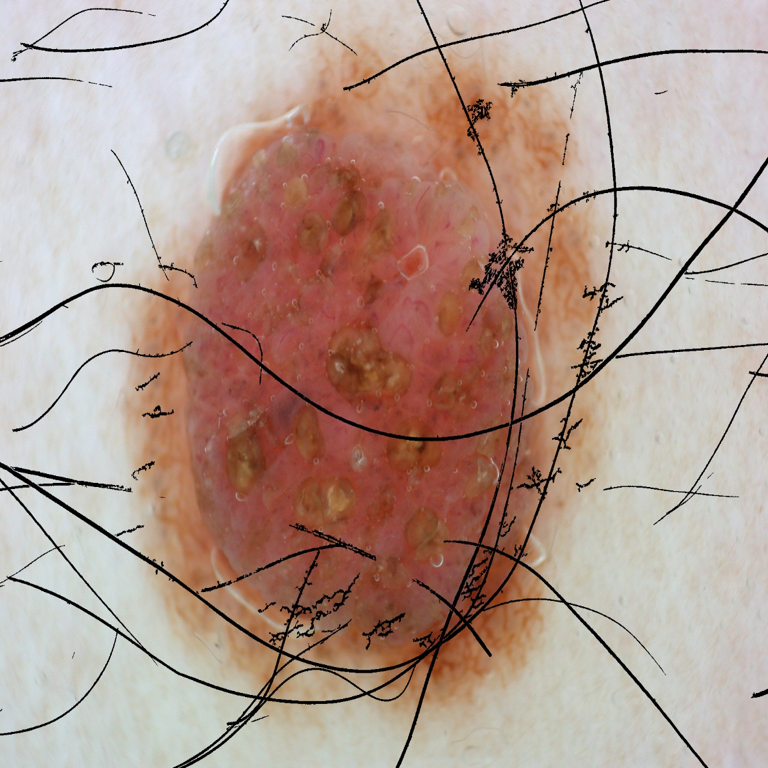}
         \caption{}
         \label{fig:ablationLoss_hair}
     \end{subfigure}
     \begin{subfigure}[b]{0.3\textwidth}
         \centering
         \includegraphics[width=\textwidth]{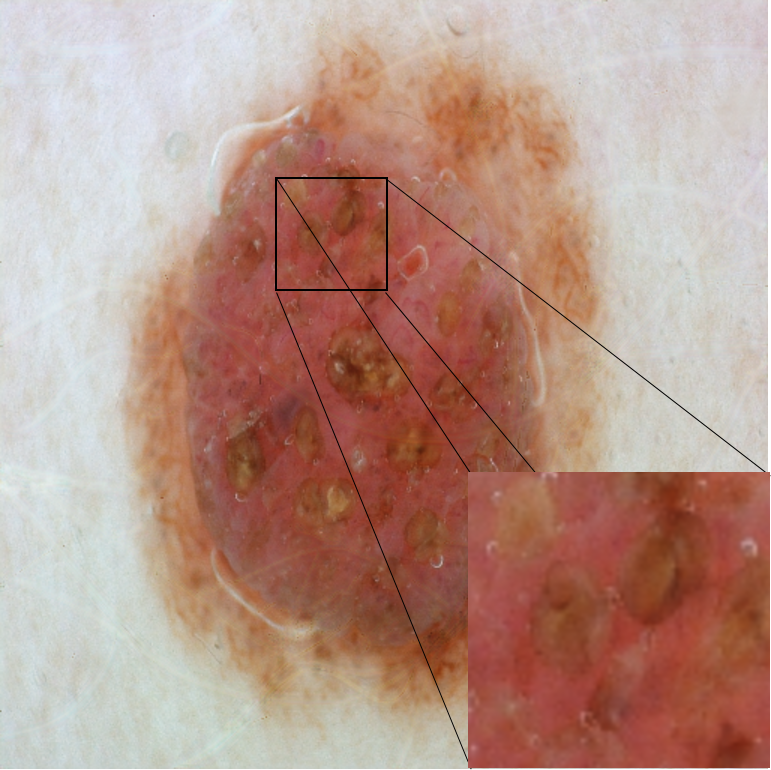}
         \caption{}
         \label{fig:ablationLoss_base}
     \end{subfigure}
     \\
     \begin{subfigure}[b]{0.3\textwidth}
         \centering
         \includegraphics[width=\textwidth]{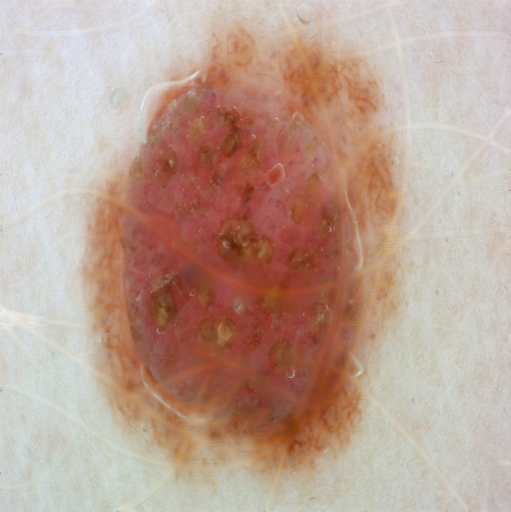}
         \caption{}
     \end{subfigure}
     \begin{subfigure}[b]{0.3\textwidth}
         \centering
         \includegraphics[width=\textwidth]{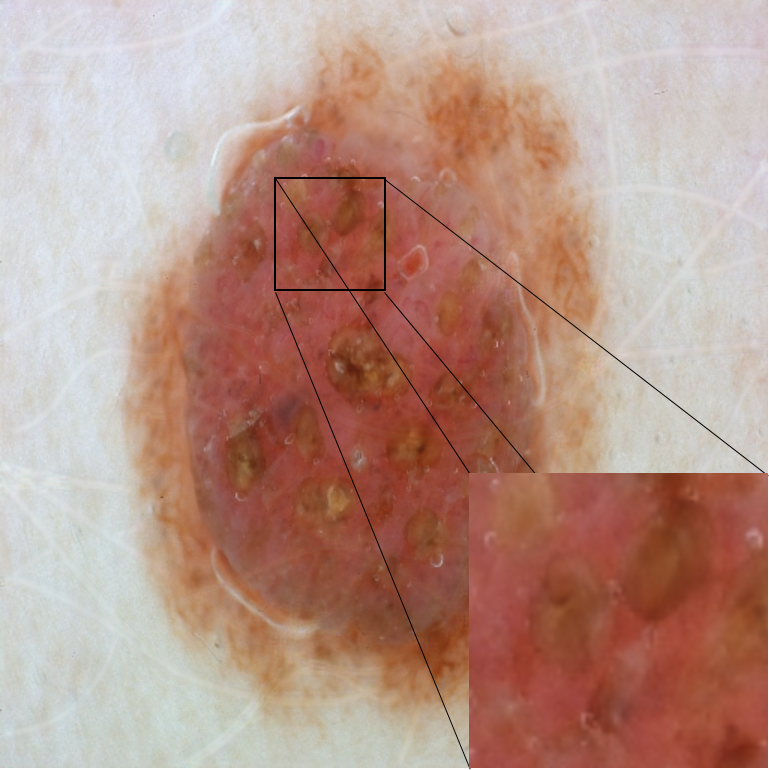}
         \caption{}
         \label{fig:ablationLoss_noBackground}
     \end{subfigure}
          \begin{subfigure}[b]{0.3\textwidth}
         \centering
         \includegraphics[width=\textwidth]{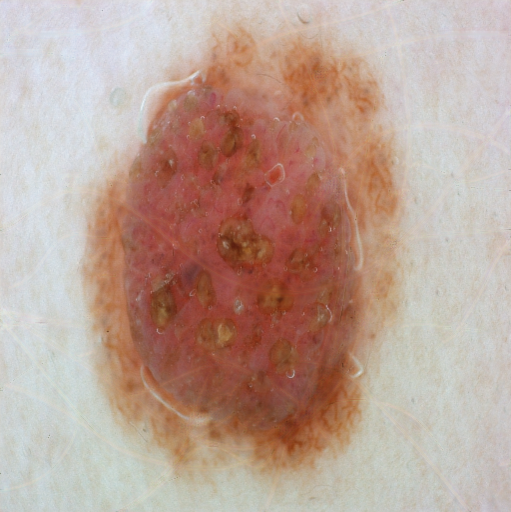}
         \caption{}
     \end{subfigure}
     \\
     \begin{subfigure}[b]{0.3\textwidth}
         \centering
         \includegraphics[width=\textwidth]{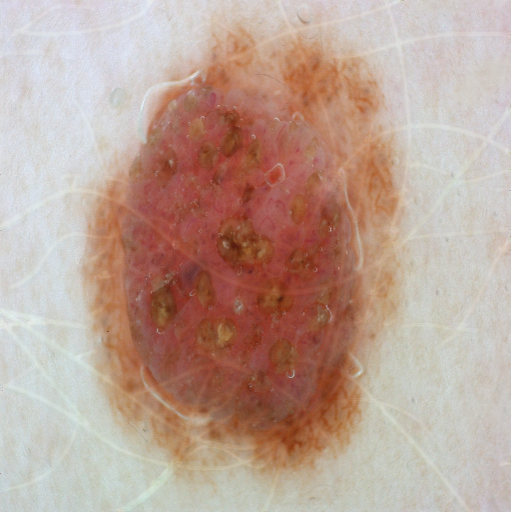}
         \caption{}
     \end{subfigure}
          \begin{subfigure}[b]{0.3\textwidth}
         \centering
         \includegraphics[width=\textwidth]{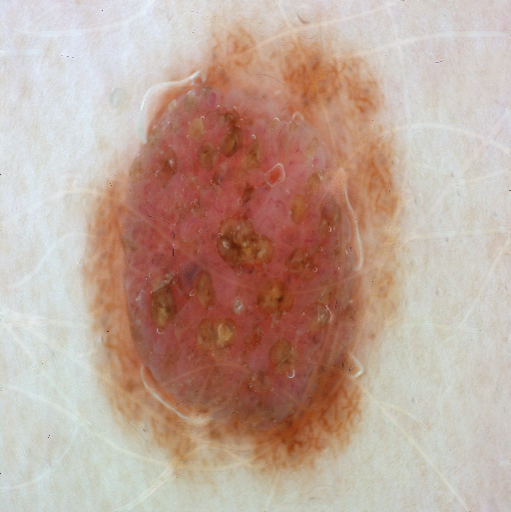}
         \caption{}
     \end{subfigure}
    \caption{(a) Original hairless image, (b) Input image with simulated hair, (c) result of our model trained with the complete loss, (d)-(h) results of our model trained by removing (d) $L_1^\text{foreground}$, (e) $L_1^\text{background}$, (f) $L_2^\text{composed}$ , (g) $\loss{SSIM}$, and (h) $\loss{tv}$.}
    \label{fig:Loss_ablation}
\end{figure}

Another study that we believe is of great importance is the evaluation of the relevance of each term of the loss function. As in the previous case, we show in Table \ref{tab:DeConvSkipvs.PoolNoSkip} and Figure \ref{fig:Loss_ablation} the quantitative and qualitative results, respectively, of our model trained by removing in each case one of the terms that compose the loss function. As can be seen, most of the performance measures and resulting images are worsen by deleting some of the terms. This is not the case when we stop computing the $L_1$ distance between the GT and the network's prediction only among the background pixels ($L_1^\text{background}$). By comparing Figures \ref{fig:ablationLoss_base} and \ref{fig:ablationLoss_noBackground}, we can see that when we do not use this term, the structures tend to be more blurrier. Such blurrier regions may not be penalized as much when calculating performance measures.

\section{Discussion and Conclusions}\label{sec:discussionresults}

In this work, we have presented a novel CNN-based method to the task of hair removal in dermoscopic images. We have built an encoder-decoder architecture, which has shown good results in reconstruction tasks like the one at hand. We highlight an architectural aspect of the network: the use of skip connections helps to retrieve details. The benefits of its use have been demonstrated with an ablation study. In addition, we have analyzed the performance of our method and compared it with six state-of-the-art approaches. To carry out the experiments we created a dataset using different hair simulating strategies over images from publicly available dermoscopic datasets. For the validation of the algorithms, we calculated nine measures of similarity between the hairless reference images and their corresponding image with simulated hair. Finally, we performed a statistical test to objectively study and compare their performance.

The results obtained by means of the statistical tests applied to these measures lead to the conclusion that for eight of the performance measures, our method is statistically the best algorithm.  Except for the VIF measure and when we compare it with Abbas' \etal and Lee \etal's methods. As reflected in Figure \ref{fig:graphic_comparation} and in Table \ref{tab:statistic_test}, Abbas' and Toossi's algorithms produce the least suitable results. This bad behavior may be due to the fact that these algorithms do not seem to distinguish well hairs of greater thickness or dark colors.

It is worth mentioning that we have evaluated our model in dermoscopic images with real hair, obtaining good visual results and demonstrating, thus, its effectiveness.

As future work, we aim to use our approach on a more complete skin lesion analysis system, leveraging the knowledge to extract other characteristics. Also, increasing the number of images used in the dataset to train the network might enhance the network's generalization capabilities.

\section*{Acknowledgments}
This work was partially supported by the project \mbox{TIN 2016-75404-P} AEI/FEDER, UE. Lidia Talavera-Mart\'inez also benefited from the fellowship \mbox{BES-2017-081264} conceded by the \emph{Ministry of Economy, Industry and Competitiveness} under a program co-financed by the \emph{European Social Fund}. We thank Dr. Mohamed Attia from the  Institute For Innovation and Research, Deakin University, Australia, for providing the GAN-based simulated hair images.

\bibliographystyle{model2-names.bst}
\bibliography{references}

\begin{thebibliography}{43}
\expandafter\ifx\csname natexlab\endcsname\relax\def\natexlab#1{#1}\fi
\providecommand{\url}[1]{\texttt{#1}}
\providecommand{\href}[2]{#2}
\providecommand{\path}[1]{#1}
\providecommand{\DOIprefix}{doi:}
\providecommand{\ArXivprefix}{arXiv:}
\providecommand{\URLprefix}{URL: }
\providecommand{\Pubmedprefix}{pmid:}
\providecommand{\doi}[1]{\href{http://dx.doi.org/#1}{\path{#1}}}
\providecommand{\Pubmed}[1]{\href{pmid:#1}{\path{#1}}}
\providecommand{\bibinfo}[2]{#2}
\ifx\xfnm\relax \def\xfnm[#1]{\unskip,\space#1}\fi
%Type = Misc
\bibitem[{ECI()}]{ECIS}
, .
\newblock \bibinfo{title}{European {C}ancer {I}nformation {S}ystem}.
\newblock
  \bibinfo{howpublished}{\url{https://ecis.jrc.ec.europa.eu/index.php}}.
\newblock \bibinfo{note}{Accessed on 4th April 2020}.
%Type = Misc
\bibitem[{mmm()}]{mmmp}
, .
\newblock \bibinfo{title}{Melanoma {M}olecular {M}ap {P}roject}.
\newblock \bibinfo{howpublished}{\url{http://www.mmmp.org/MMMP/welcome.mmmp}}.
\newblock \bibinfo{note}{Accessed on 4th April 2020}.
%Type = Article
\bibitem[{Abbas et~al.(2011)Abbas, Celebi and Garc{\'\i}a}]{abbas2011hair}
\bibinfo{author}{Abbas, Q.}, \bibinfo{author}{Celebi, M.E.},
  \bibinfo{author}{Garc{\'\i}a, I.F.}, \bibinfo{year}{2011}.
\newblock \bibinfo{title}{Hair removal methods: a comparative study for
  dermoscopy images}.
\newblock \bibinfo{journal}{Biomedical Signal Processing and Control}
  \bibinfo{volume}{6}, \bibinfo{pages}{395--404}.
%Type = Article
\bibitem[{Argenziano et~al.(1998)Argenziano, Fabbrocini, Carli, De~Giorgi,
  Sammarco and Delfino}]{argenziano1998epiluminescence}
\bibinfo{author}{Argenziano, G.}, \bibinfo{author}{Fabbrocini, G.},
  \bibinfo{author}{Carli, P.}, \bibinfo{author}{De~Giorgi, V.},
  \bibinfo{author}{Sammarco, E.}, \bibinfo{author}{Delfino, M.},
  \bibinfo{year}{1998}.
\newblock \bibinfo{title}{Epiluminescence microscopy for the diagnosis of
  doubtful melanocytic skin lesions: comparison of the {ABCD} rule of
  dermatoscopy and a new 7-point checklist based on pattern analysis}.
\newblock \bibinfo{journal}{Archives of dermatology} \bibinfo{volume}{134},
  \bibinfo{pages}{1563--1570}.
%Type = Book
\bibitem[{Argenziano et~al.(2000)Argenziano, Soyer, De~Giorgi, Piccolo, Carli
  and Delfino}]{argenziano2000interactive}
\bibinfo{author}{Argenziano, G.}, \bibinfo{author}{Soyer, H.},
  \bibinfo{author}{De~Giorgi, V.}, \bibinfo{author}{Piccolo, D.},
  \bibinfo{author}{Carli, P.}, \bibinfo{author}{Delfino, M.},
  \bibinfo{year}{2000}.
\newblock \bibinfo{title}{Interactive atlas of dermoscopy (Book and CD-ROM)}.
\newblock \bibinfo{publisher}{{EDRA} {M}edical {P}ublishing \& {N}ew media}.
%Type = Article
\bibitem[{Argenziano et~al.(2003)Argenziano, Soyer, Chimenti, Talamini, Corona,
  Sera, Binder, Cerroni, De~Rosa, Ferrara et~al.}]{argenziano2003dermoscopy}
\bibinfo{author}{Argenziano, G.}, \bibinfo{author}{Soyer, H.P.},
  \bibinfo{author}{Chimenti, S.}, \bibinfo{author}{Talamini, R.},
  \bibinfo{author}{Corona, R.}, \bibinfo{author}{Sera, F.},
  \bibinfo{author}{Binder, M.}, \bibinfo{author}{Cerroni, L.},
  \bibinfo{author}{De~Rosa, G.}, \bibinfo{author}{Ferrara, G.}, et~al.,
  \bibinfo{year}{2003}.
\newblock \bibinfo{title}{Dermoscopy of pigmented skin lesions: results of a
  consensus meeting via the internet}.
\newblock \bibinfo{journal}{Journal of the American Academy of Dermatology}
  \bibinfo{volume}{48}, \bibinfo{pages}{679--693}.
%Type = Article
\bibitem[{Attia et~al.(2020)Attia, Hossny, Zhou, Nahavandi, Asadi and
  Yazdabadi}]{attia2020realistic}
\bibinfo{author}{Attia, M.}, \bibinfo{author}{Hossny, M.},
  \bibinfo{author}{Zhou, H.}, \bibinfo{author}{Nahavandi, S.},
  \bibinfo{author}{Asadi, H.}, \bibinfo{author}{Yazdabadi, A.},
  \bibinfo{year}{2020}.
\newblock \bibinfo{title}{Realistic hair simulator for skin lesion images: A
  novel benchemarking tool}.
\newblock \bibinfo{journal}{Artificial Intelligence in Medicine}
  \bibinfo{volume}{108}, \bibinfo{pages}{101933}.
%Type = Article
\bibitem[{Bakkouri and Afdel(2020)}]{bakkouri2020computer}
\bibinfo{author}{Bakkouri, I.}, \bibinfo{author}{Afdel, K.},
  \bibinfo{year}{2020}.
\newblock \bibinfo{title}{Computer-aided diagnosis (cad) system based on
  multi-layer feature fusion network for skin lesion recognition in dermoscopy
  images}.
\newblock \bibinfo{journal}{Multimedia Tools and Applications}
  \bibinfo{volume}{79}, \bibinfo{pages}{20483--20518}.
%Type = Article
\bibitem[{Barnston(1992)}]{barnston1992correspondence}
\bibinfo{author}{Barnston, A.G.}, \bibinfo{year}{1992}.
\newblock \bibinfo{title}{Correspondence among the correlation, {RMSE}, and
  {H}eidke forecast verification measures; refinement of the {H}eidke score}.
\newblock \bibinfo{journal}{Weather and Forecasting} \bibinfo{volume}{7},
  \bibinfo{pages}{699--709}.
%Type = Inproceedings
\bibitem[{Bibiloni et~al.(2017)Bibiloni, Gonz{\'a}lez-Hidalgo and
  Massanet}]{bibiloni2017skin}
\bibinfo{author}{Bibiloni, P.}, \bibinfo{author}{Gonz{\'a}lez-Hidalgo, M.},
  \bibinfo{author}{Massanet, S.}, \bibinfo{year}{2017}.
\newblock \bibinfo{title}{Skin hair removal in dermoscopic images using soft
  color morphology}, in: \bibinfo{booktitle}{Conference on Artificial
  Intelligence in Medicine in Europe}, \bibinfo{organization}{Springer}. pp.
  \bibinfo{pages}{322--326}.
%Type = Article
\bibitem[{Chollet et~al.(2018)}]{chollet2018keras}
\bibinfo{author}{Chollet, F.}, et~al., \bibinfo{year}{2018}.
\newblock \bibinfo{title}{Keras: The python deep learning library}.
\newblock \bibinfo{journal}{Astrophysics Source Code Library} .
%Type = Inproceedings
\bibitem[{Cui et~al.(2014)Cui, Chang, Shan, Zhong and Chen}]{cui2014deep}
\bibinfo{author}{Cui, Z.}, \bibinfo{author}{Chang, H.}, \bibinfo{author}{Shan,
  S.}, \bibinfo{author}{Zhong, B.}, \bibinfo{author}{Chen, X.},
  \bibinfo{year}{2014}.
\newblock \bibinfo{title}{Deep network cascade for image super-resolution}, in:
  \bibinfo{booktitle}{European Conference on Computer Vision},
  \bibinfo{organization}{Springer}. pp. \bibinfo{pages}{49--64}.
%Type = Article
\bibitem[{Dong et~al.(2015)Dong, Loy, He and Tang}]{dong2015image}
\bibinfo{author}{Dong, C.}, \bibinfo{author}{Loy, C.C.}, \bibinfo{author}{He,
  K.}, \bibinfo{author}{Tang, X.}, \bibinfo{year}{2015}.
\newblock \bibinfo{title}{Image super-resolution using deep convolutional
  networks}.
\newblock \bibinfo{journal}{IEEE transactions on pattern analysis and machine
  intelligence} \bibinfo{volume}{38}, \bibinfo{pages}{295--307}.
%Type = Inproceedings
\bibitem[{Egiazarian et~al.(2006)Egiazarian, Astola, Ponomarenko, Lukin,
  Battisti and Carli}]{egiazarian2006new}
\bibinfo{author}{Egiazarian, K.}, \bibinfo{author}{Astola, J.},
  \bibinfo{author}{Ponomarenko, N.}, \bibinfo{author}{Lukin, V.},
  \bibinfo{author}{Battisti, F.}, \bibinfo{author}{Carli, M.},
  \bibinfo{year}{2006}.
\newblock \bibinfo{title}{New full-reference quality metrics based on hvs}, in:
  \bibinfo{booktitle}{Proceedings of the Second International Workshop on Video
  Processing and Quality Metrics}.
%Type = Inproceedings
\bibitem[{Huang et~al.(2013)Huang, Kwan, Chang, Liu, Chi and
  Chen}]{huang2013robust}
\bibinfo{author}{Huang, A.}, \bibinfo{author}{Kwan, S.Y.},
  \bibinfo{author}{Chang, W.Y.}, \bibinfo{author}{Liu, M.Y.},
  \bibinfo{author}{Chi, M.H.}, \bibinfo{author}{Chen, G.S.},
  \bibinfo{year}{2013}.
\newblock \bibinfo{title}{A robust hair segmentation and removal approach for
  clinical images of skin lesions}, in: \bibinfo{booktitle}{2013 35th Annual
  International Conference of the IEEE Engineering in Medicine and Biology
  Society (EMBC)}, \bibinfo{organization}{IEEE}. pp.
  \bibinfo{pages}{3315--3318}.
%Type = Inproceedings
\bibitem[{Jain and Seung(2009)}]{jain2009natural}
\bibinfo{author}{Jain, V.}, \bibinfo{author}{Seung, S.}, \bibinfo{year}{2009}.
\newblock \bibinfo{title}{Natural image denoising with convolutional networks},
  in: \bibinfo{booktitle}{Advances in neural information processing systems},
  pp. \bibinfo{pages}{769--776}.
%Type = Article
\bibitem[{Kingma and Ba(2014)}]{kingma2014adam}
\bibinfo{author}{Kingma, D.P.}, \bibinfo{author}{Ba, J.}, \bibinfo{year}{2014}.
\newblock \bibinfo{title}{Adam: A method for stochastic optimization}.
\newblock \bibinfo{journal}{arXiv preprint arXiv:1412.6980} .
%Type = Article
\bibitem[{Kittler(2007)}]{kittler2007dermatoscopy}
\bibinfo{author}{Kittler, H.}, \bibinfo{year}{2007}.
\newblock \bibinfo{title}{Dermatoscopy: introduction of a new algorithmic
  method based on pattern analysis for diagnosis of pigmented skin lesions}.
\newblock \bibinfo{journal}{Dermatopathology: Practical \& Conceptual}
  \bibinfo{volume}{13}, \bibinfo{pages}{3}.
%Type = Article
\bibitem[{Lee et~al.(1997)Lee, Ng, Gallagher, Coldman and
  McLean}]{lee1997dullrazor}
\bibinfo{author}{Lee, T.}, \bibinfo{author}{Ng, V.},
  \bibinfo{author}{Gallagher, R.}, \bibinfo{author}{Coldman, A.},
  \bibinfo{author}{McLean, D.}, \bibinfo{year}{1997}.
\newblock \bibinfo{title}{Dullrazor{\textregistered}: A software approach to
  hair removal from images}.
\newblock \bibinfo{journal}{Computers in biology and medicine}
  \bibinfo{volume}{27}, \bibinfo{pages}{533--543}.
%Type = Article
\bibitem[{Litjens et~al.(2017)Litjens, Kooi, Bejnordi, Setio, Ciompi,
  Ghafoorian, Van Der~Laak, Van~Ginneken and S{\'a}nchez}]{litjens2017survey}
\bibinfo{author}{Litjens, G.}, \bibinfo{author}{Kooi, T.},
  \bibinfo{author}{Bejnordi, B.E.}, \bibinfo{author}{Setio, A.A.A.},
  \bibinfo{author}{Ciompi, F.}, \bibinfo{author}{Ghafoorian, M.},
  \bibinfo{author}{Van Der~Laak, J.A.}, \bibinfo{author}{Van~Ginneken, B.},
  \bibinfo{author}{S{\'a}nchez, C.I.}, \bibinfo{year}{2017}.
\newblock \bibinfo{title}{A survey on deep learning in medical image analysis}.
\newblock \bibinfo{journal}{Medical image analysis} \bibinfo{volume}{42},
  \bibinfo{pages}{60--88}.
%Type = Inproceedings
\bibitem[{Liu et~al.(2018)Liu, Reda, Shih, Wang, Tao and
  Catanzaro}]{liu2018image}
\bibinfo{author}{Liu, G.}, \bibinfo{author}{Reda, F.A.}, \bibinfo{author}{Shih,
  K.J.}, \bibinfo{author}{Wang, T.C.}, \bibinfo{author}{Tao, A.},
  \bibinfo{author}{Catanzaro, B.}, \bibinfo{year}{2018}.
\newblock \bibinfo{title}{Image inpainting for irregular holes using partial
  convolutions}, in: \bibinfo{booktitle}{Proceedings of the European Conference
  on Computer Vision (ECCV)}, pp. \bibinfo{pages}{85--100}.
%Type = Inproceedings
\bibitem[{Mao et~al.(2016)Mao, Shen and Yang}]{mao2016image}
\bibinfo{author}{Mao, X.}, \bibinfo{author}{Shen, C.}, \bibinfo{author}{Yang,
  Y.B.}, \bibinfo{year}{2016}.
\newblock \bibinfo{title}{Image restoration using very deep convolutional
  encoder-decoder networks with symmetric skip connections}, in:
  \bibinfo{booktitle}{Advances in neural information processing systems}, pp.
  \bibinfo{pages}{2802--2810}.
%Type = Article
\bibitem[{Mayer et~al.(1997)}]{mayer1997systematic}
\bibinfo{author}{Mayer, J.}, et~al., \bibinfo{year}{1997}.
\newblock \bibinfo{title}{Systematic review of the diagnostic accuracy of
  dermatoscopy in detecting malignant melanoma}.
\newblock \bibinfo{journal}{Medical journal of Australia}
  \bibinfo{volume}{167}, \bibinfo{pages}{206--210}.
%Type = Article
\bibitem[{Mendon{\c{c}}a et~al.(2015)Mendon{\c{c}}a, Celebi, Mendon{\c{c}}a and
  Marques}]{mendonca2015ph2}
\bibinfo{author}{Mendon{\c{c}}a, T.}, \bibinfo{author}{Celebi, M.},
  \bibinfo{author}{Mendon{\c{c}}a, T.}, \bibinfo{author}{Marques, J.},
  \bibinfo{year}{2015}.
\newblock \bibinfo{title}{${PH}^2$: A public database for the analysis of
  dermoscopic images}.
\newblock \bibinfo{journal}{Dermoscopy Image Analysis} .
%Type = Article
\bibitem[{Menzies et~al.(1996)Menzies, Ingvar, Crotty and
  McCarthy}]{menzies1996frequency}
\bibinfo{author}{Menzies, S.W.}, \bibinfo{author}{Ingvar, C.},
  \bibinfo{author}{Crotty, K.A.}, \bibinfo{author}{McCarthy, W.H.},
  \bibinfo{year}{1996}.
\newblock \bibinfo{title}{Frequency and morphologic characteristics of invasive
  melanomas lacking specific surface microscopic features}.
\newblock \bibinfo{journal}{Archives of Dermatology} \bibinfo{volume}{132},
  \bibinfo{pages}{1178--1182}.
%Type = Misc
\bibitem[{{M}irzaalian()}]{hairsim}
\bibinfo{author}{{M}irzaalian, H.}, .
\newblock \bibinfo{title}{Hair {S}im {S}oftware}.
\newblock
  \bibinfo{howpublished}{\url{http://www2.cs.sfu.ca/~hamarneh/software/hairsim/Welcome.html}}.
\newblock \bibinfo{note}{Accessed on 7th Mar 2019}.
%Type = Article
\bibitem[{Mirzaalian et~al.(2014)Mirzaalian, Lee and
  Hamarneh}]{mirzaalian2014hair}
\bibinfo{author}{Mirzaalian, H.}, \bibinfo{author}{Lee, T.K.},
  \bibinfo{author}{Hamarneh, G.}, \bibinfo{year}{2014}.
\newblock \bibinfo{title}{Hair enhancement in dermoscopic images using
  dual-channel quaternion tubularness filters and {MRF}-based multilabel
  optimization}.
\newblock \bibinfo{journal}{IEEE Transactions on Image Processing}
  \bibinfo{volume}{23}, \bibinfo{pages}{5486--5496}.
%Type = Inproceedings
\bibitem[{O’Mahony et~al.(2019)O’Mahony, Campbell, Carvalho, Harapanahalli,
  Hernandez, Krpalkova, Riordan and Walsh}]{o2019deep}
\bibinfo{author}{O’Mahony, N.}, \bibinfo{author}{Campbell, S.},
  \bibinfo{author}{Carvalho, A.}, \bibinfo{author}{Harapanahalli, S.},
  \bibinfo{author}{Hernandez, G.V.}, \bibinfo{author}{Krpalkova, L.},
  \bibinfo{author}{Riordan, D.}, \bibinfo{author}{Walsh, J.},
  \bibinfo{year}{2019}.
\newblock \bibinfo{title}{Deep learning vs. traditional computer vision}, in:
  \bibinfo{booktitle}{Science and Information Conference},
  \bibinfo{organization}{Springer}. pp. \bibinfo{pages}{128--144}.
%Type = Article
\bibitem[{Rudin et~al.(1992)Rudin, Osher and Fatemi}]{rudin1992nonlinear}
\bibinfo{author}{Rudin, L.I.}, \bibinfo{author}{Osher, S.},
  \bibinfo{author}{Fatemi, E.}, \bibinfo{year}{1992}.
\newblock \bibinfo{title}{Nonlinear total variation based noise removal
  algorithms}.
\newblock \bibinfo{journal}{Physica D: nonlinear phenomena}
  \bibinfo{volume}{60}, \bibinfo{pages}{259--268}.
%Type = Article
\bibitem[{Salido and Ruiz(2018)}]{salido2018using}
\bibinfo{author}{Salido, J.A.A.}, \bibinfo{author}{Ruiz, C.},
  \bibinfo{year}{2018}.
\newblock \bibinfo{title}{Using deep learning to detect melanoma in dermoscopy
  images}.
\newblock \bibinfo{journal}{International Journal of Machine Learning and
  Computing} \bibinfo{volume}{8}.
%Type = Inproceedings
\bibitem[{Sheikh and Bovik(2004)}]{sheikh2004image}
\bibinfo{author}{Sheikh, H.R.}, \bibinfo{author}{Bovik, A.C.},
  \bibinfo{year}{2004}.
\newblock \bibinfo{title}{Image information and visual quality}, in:
  \bibinfo{booktitle}{2004 IEEE International Conference on Acoustics, Speech,
  and Signal Processing}, \bibinfo{organization}{IEEE}. pp.
  \bibinfo{pages}{iii--709}.
%Type = Article
\bibitem[{Stolz(1994)}]{stolz1994abcd}
\bibinfo{author}{Stolz, W.}, \bibinfo{year}{1994}.
\newblock \bibinfo{title}{{ABCD} rule of dermatoscopy: a new practical method
  for early recognition of malignant melanoma}.
\newblock \bibinfo{journal}{Eur. J. Dermatol.} \bibinfo{volume}{4},
  \bibinfo{pages}{521--527}.
%Type = Inproceedings
\bibitem[{Talavera-Mart{\'\i}nez et~al.(2019)Talavera-Mart{\'\i}nez, Bibiloni
  and Gonz{\'a}lez-Hidalgo}]{talavera2019comparative}
\bibinfo{author}{Talavera-Mart{\'\i}nez, L.}, \bibinfo{author}{Bibiloni, P.},
  \bibinfo{author}{Gonz{\'a}lez-Hidalgo, M.}, \bibinfo{year}{2019}.
\newblock \bibinfo{title}{Comparative study of dermoscopic hair removal
  methods}, in: \bibinfo{booktitle}{ECCOMAS Thematic Conference on
  Computational Vision and Medical Image Processing},
  \bibinfo{organization}{Springer}. pp. \bibinfo{pages}{12--21}.
%Type = Article
\bibitem[{Tian et~al.(2019)Tian, Fei, Zheng, Xu, Zuo and Lin}]{tian2019deep}
\bibinfo{author}{Tian, C.}, \bibinfo{author}{Fei, L.}, \bibinfo{author}{Zheng,
  W.}, \bibinfo{author}{Xu, Y.}, \bibinfo{author}{Zuo, W.},
  \bibinfo{author}{Lin, C.W.}, \bibinfo{year}{2019}.
\newblock \bibinfo{title}{Deep learning on image denoising: An overview}.
\newblock \bibinfo{journal}{arXiv preprint arXiv:1912.13171} .
%Type = Article
\bibitem[{Toossi et~al.(2013)Toossi, Pourreza, Zare, Sigari, Layegh and
  Azimi}]{toossi2013effective}
\bibinfo{author}{Toossi, M.T.B.}, \bibinfo{author}{Pourreza, H.R.},
  \bibinfo{author}{Zare, H.}, \bibinfo{author}{Sigari, M.H.},
  \bibinfo{author}{Layegh, P.}, \bibinfo{author}{Azimi, A.},
  \bibinfo{year}{2013}.
\newblock \bibinfo{title}{An effective hair removal algorithm for dermoscopy
  images}.
\newblock \bibinfo{journal}{Skin Research and Technology} \bibinfo{volume}{19},
  \bibinfo{pages}{230--235}.
%Type = Inproceedings
\bibitem[{Vincent et~al.(2008)Vincent, Larochelle, Bengio and
  Manzagol}]{vincent2008extracting}
\bibinfo{author}{Vincent, P.}, \bibinfo{author}{Larochelle, H.},
  \bibinfo{author}{Bengio, Y.}, \bibinfo{author}{Manzagol, P.A.},
  \bibinfo{year}{2008}.
\newblock \bibinfo{title}{Extracting and composing robust features with
  denoising autoencoders}, in: \bibinfo{booktitle}{Proceedings of the 25th
  international conference on Machine learning}, pp.
  \bibinfo{pages}{1096--1103}.
%Type = Article
\bibitem[{Wang and Bovik(2002)}]{wang2002universal}
\bibinfo{author}{Wang, Z.}, \bibinfo{author}{Bovik, A.C.},
  \bibinfo{year}{2002}.
\newblock \bibinfo{title}{A universal image quality index}.
\newblock \bibinfo{journal}{IEEE signal processing letters}
  \bibinfo{volume}{9}, \bibinfo{pages}{81--84}.
%Type = Article
\bibitem[{Wang and Bovik(2009)}]{wang2009mean}
\bibinfo{author}{Wang, Z.}, \bibinfo{author}{Bovik, A.C.},
  \bibinfo{year}{2009}.
\newblock \bibinfo{title}{Mean squared error: Love it or leave it? a new look
  at signal fidelity measures}.
\newblock \bibinfo{journal}{IEEE signal processing magazine}
  \bibinfo{volume}{26}, \bibinfo{pages}{98--117}.
%Type = Article
\bibitem[{Wang et~al.(2004)Wang, Bovik, Sheikh, Simoncelli
  et~al.}]{wang2004image}
\bibinfo{author}{Wang, Z.}, \bibinfo{author}{Bovik, A.C.},
  \bibinfo{author}{Sheikh, H.R.}, \bibinfo{author}{Simoncelli, E.P.}, et~al.,
  \bibinfo{year}{2004}.
\newblock \bibinfo{title}{Image quality assessment: from error visibility to
  structural similarity}.
\newblock \bibinfo{journal}{IEEE transactions on image processing}
  \bibinfo{volume}{13}, \bibinfo{pages}{600--612}.
%Type = Inproceedings
\bibitem[{Wang et~al.(2003)Wang, Simoncelli and Bovik}]{wang2003multiscale}
\bibinfo{author}{Wang, Z.}, \bibinfo{author}{Simoncelli, E.P.},
  \bibinfo{author}{Bovik, A.C.}, \bibinfo{year}{2003}.
\newblock \bibinfo{title}{Multiscale structural similarity for image quality
  assessment}, in: \bibinfo{booktitle}{The Thrity-Seventh Asilomar Conference
  on Signals, Systems \& Computers, 2003}, \bibinfo{organization}{Ieee}. pp.
  \bibinfo{pages}{1398--1402}.
%Type = Article
\bibitem[{Xie et~al.(2009)Xie, Qin, Jiang and Meng}]{xie2009pde}
\bibinfo{author}{Xie, F.Y.}, \bibinfo{author}{Qin, S.Y.},
  \bibinfo{author}{Jiang, Z.G.}, \bibinfo{author}{Meng, R.S.},
  \bibinfo{year}{2009}.
\newblock \bibinfo{title}{{PDE}-based unsupervised repair of hair-occluded
  information in dermoscopy images of melanoma}.
\newblock \bibinfo{journal}{Computerized Medical Imaging and Graphics}
  \bibinfo{volume}{33}, \bibinfo{pages}{275--282}.
%Type = Inproceedings
\bibitem[{Xie et~al.(2012)Xie, Xu and Chen}]{xie2012image}
\bibinfo{author}{Xie, J.}, \bibinfo{author}{Xu, L.}, \bibinfo{author}{Chen,
  E.}, \bibinfo{year}{2012}.
\newblock \bibinfo{title}{Image denoising and inpainting with deep neural
  networks}, in: \bibinfo{booktitle}{Advances in neural information processing
  systems}, pp. \bibinfo{pages}{341--349}.
%Type = Article
\bibitem[{Zhao et~al.(2016)Zhao, Gallo, Frosio and Kautz}]{zhao2016loss}
\bibinfo{author}{Zhao, H.}, \bibinfo{author}{Gallo, O.},
  \bibinfo{author}{Frosio, I.}, \bibinfo{author}{Kautz, J.},
  \bibinfo{year}{2016}.
\newblock \bibinfo{title}{Loss functions for image restoration with neural
  networks}.
\newblock \bibinfo{journal}{IEEE Transactions on computational imaging}
  \bibinfo{volume}{3}, \bibinfo{pages}{47--57}.

\end{thebibliography}
\end{document}